\documentclass[10pt,twocolumn,letterpaper]{article}

\usepackage{iccv}
\usepackage{times}
\usepackage{epsfig}
\usepackage{graphicx}
\usepackage{amsmath}
\usepackage{amssymb}

\usepackage[accsupp]{axessibility}  

\usepackage{booktabs}
\usepackage{bm}
\usepackage{subcaption}
\usepackage{enumitem}
\DeclareMathOperator*{\argmax}{argmax} 
\newcommand\norm[1]{\left\lVert#1\right\rVert}
\usepackage{authblk}
\usepackage[dvipsnames]{xcolor}
\newcommand{\red}[1]{\textcolor{red}{#1}}
\newcommand{\blue}[1]{\textcolor{blue}{#1}}
\usepackage[pagebackref=true,breaklinks=true,letterpaper=true,colorlinks,bookmarks=false]{hyperref}

\iccvfinalcopy 

\def\iccvPaperID{6024} 

\ificcvfinal\fi
\begin{document}
	
	\title{LAN-HDR: Luminance-based Alignment Network for High Dynamic Range Video Reconstruction}
	\author[1]{Haesoo Chung}
	\author[1,2]{Nam Ik Cho}
	\affil[1]{Department of Electrical and Computer Engineering, INMC, Seoul National University, Korea}
	\affil[2]{IPAI, Seoul National University, Korea}
	\affil[ ]{\tt\small \{reneeish,nicho\}@snu.ac.kr}
	\maketitle
	
	\begin{abstract}
		As demands for high-quality videos continue to rise, high-resolution and high-dynamic range (HDR) imaging techniques are drawing attention. To generate an HDR video from low dynamic range (LDR) images, one of the critical steps is the motion compensation between LDR frames, for which most existing works employed the optical flow algorithm. However, these methods suffer from flow estimation errors when saturation or complicated motions exist. In this paper, we propose an end-to-end HDR video composition framework, which aligns LDR frames in the feature space and then merges aligned features into an HDR frame, without relying on pixel-domain optical flow. Specifically, we propose a luminance-based alignment network for HDR (LAN-HDR) consisting of an alignment module and a hallucination module. The alignment module aligns a frame to the adjacent reference by evaluating luminance-based attention, excluding color information. The hallucination module generates sharp details, especially for washed-out areas due to saturation. The aligned and hallucinated features are then blended adaptively to complement each other. Finally, we merge the features to generate a final HDR frame. In training, we adopt a temporal loss, in addition to frame reconstruction losses, to enhance temporal consistency and thus reduce flickering. Extensive experiments demonstrate that our method performs better or comparable to state-of-the-art methods on several benchmarks. Codes are available at \url{https://github.com/haesoochung/LAN-HDR}.
		
	\end{abstract}
	
	\newcommand{\mcaption}{-1.3mm}
	\newcommand{\msmall}{-0.1mm}
	\newcommand{\wtext}{0.018}
	\newcommand{\wfigures}{0.48}
	\newcommand{\wteaser}{0.308}
	\newcommand{\hteaser}{1.75cm}
	\newcommand{\wteasers}{0.148}
	\newcommand{\hteasers}{1.28cm}
	\begin{figure}[t]
		\begin{minipage}{\wtext\textwidth}	
			\rotatebox{90}{\small  Inputs \quad } 
		\end{minipage}
		\begin{minipage}{\wfigures\textwidth}
			\begin{subfigure}{\wteaser\textwidth}
				\caption*{Low} 
				\includegraphics[height=\hteaser]{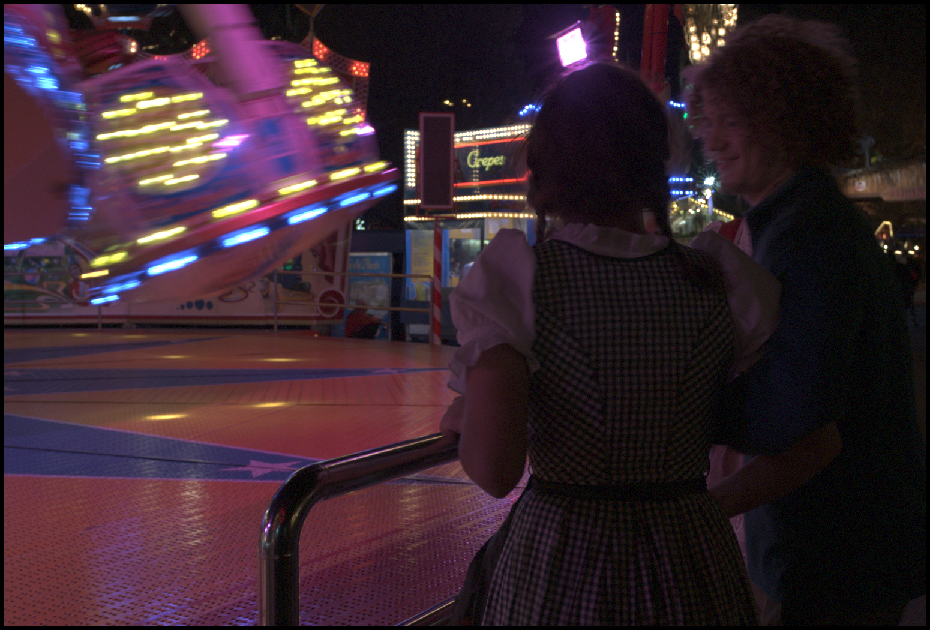} 
			\end{subfigure}	
			\begin{subfigure}{\wteaser\textwidth}
				\caption*{Middle} 
				\includegraphics[height=\hteaser]{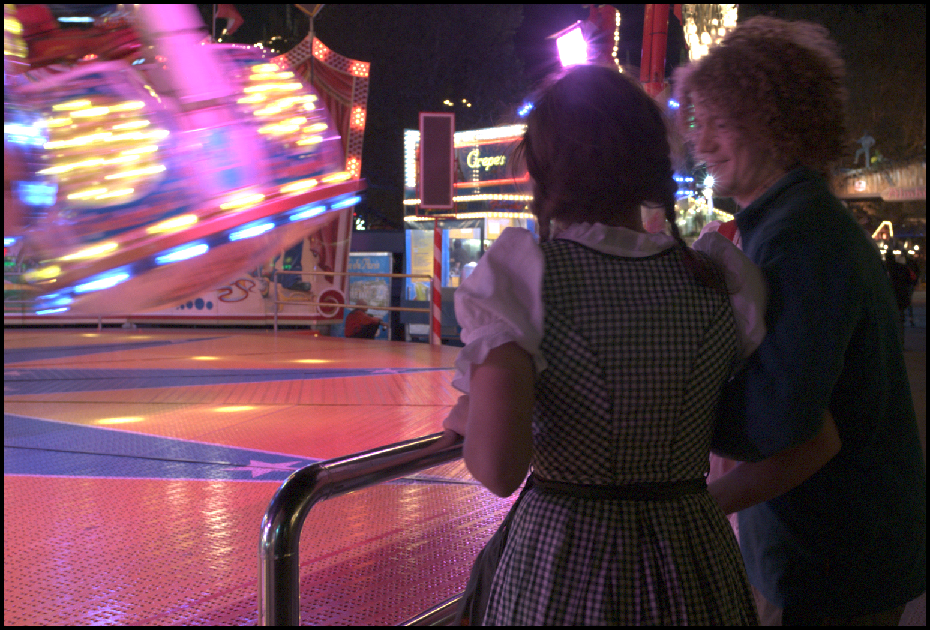}
			\end{subfigure}	
			\begin{subfigure}{\wteaser\textwidth}
				\caption*{High} 
				\includegraphics[height=\hteaser]{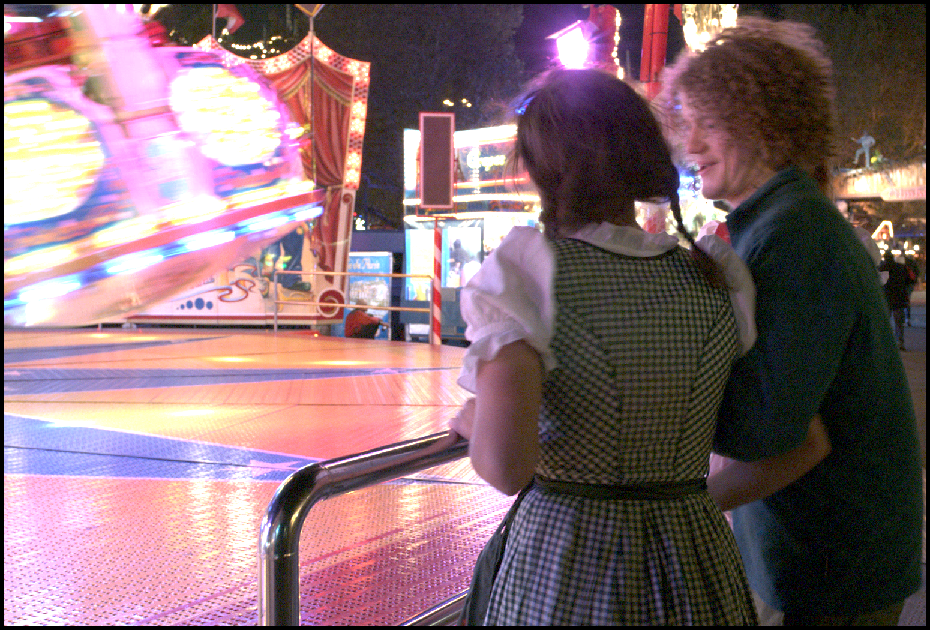}
			\end{subfigure}	
		\end{minipage}	
		\begin{minipage}{\wtext\textwidth}
			\rotatebox{90}{\small Y channels} 		
		\end{minipage}
		\begin{minipage}{\wfigures\textwidth}
			\begin{subfigure}{\wteaser\textwidth}
				\includegraphics[height=\hteaser]{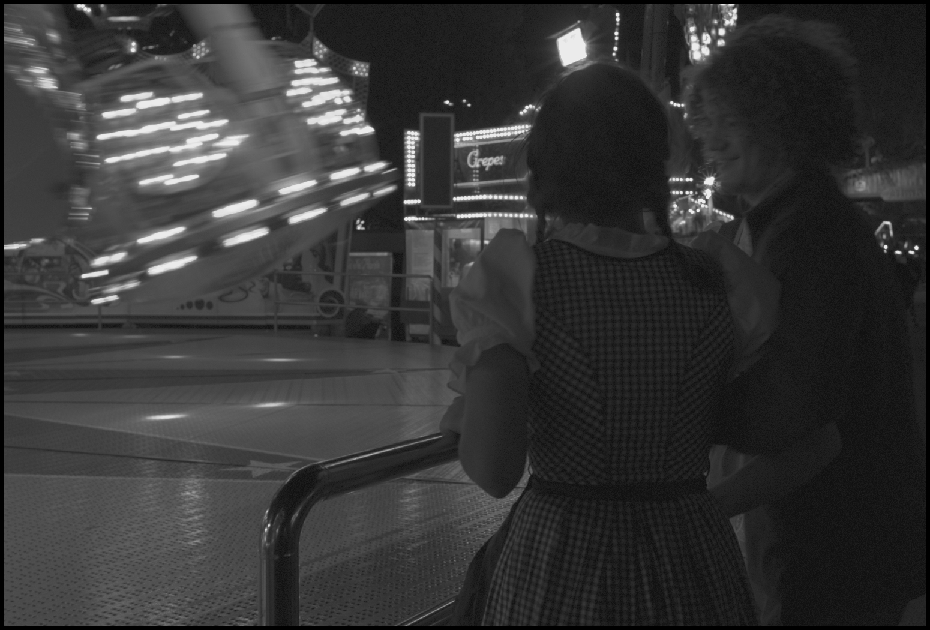}
			\end{subfigure}	
			\begin{subfigure}{\wteaser\textwidth}
				\includegraphics[height=\hteaser]{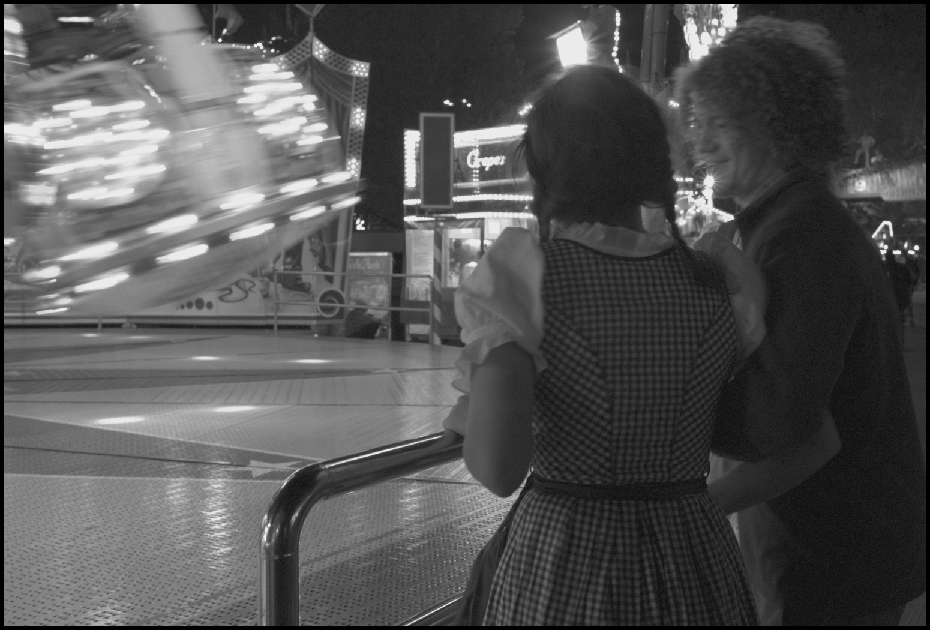}
			\end{subfigure}	
			\begin{subfigure}{\wteaser\textwidth}	
				\includegraphics[height=\hteaser]{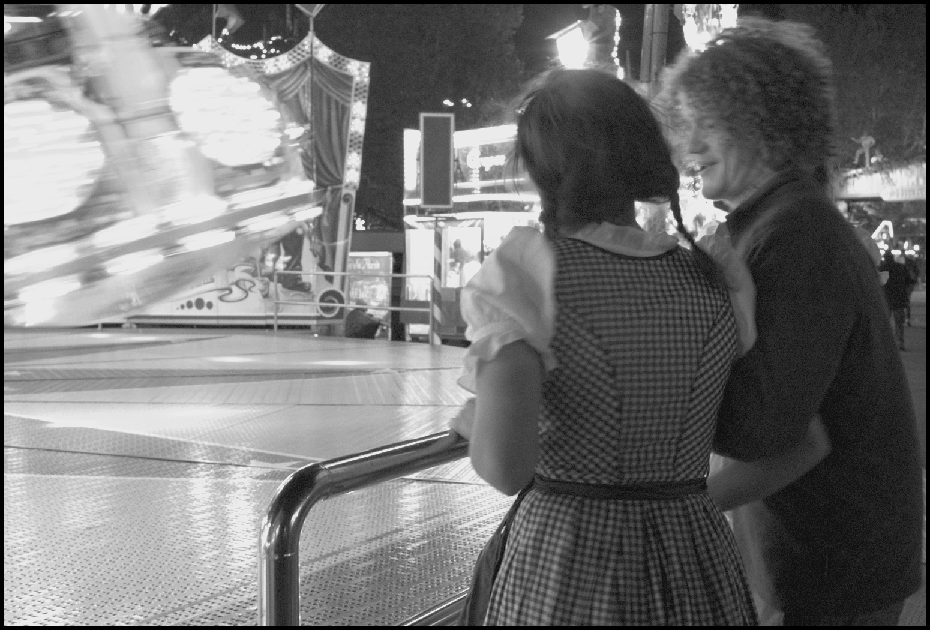}
			\end{subfigure}	
		\end{minipage}
		
		\begin{minipage}{\wtext\textwidth}
			\rotatebox{90}{\small \quad \ Our Results} 
		\end{minipage}
		\begin{minipage}{\wfigures\textwidth}	
			\begin{subfigure}{\wteaser\textwidth}
				\includegraphics[height=\hteaser]{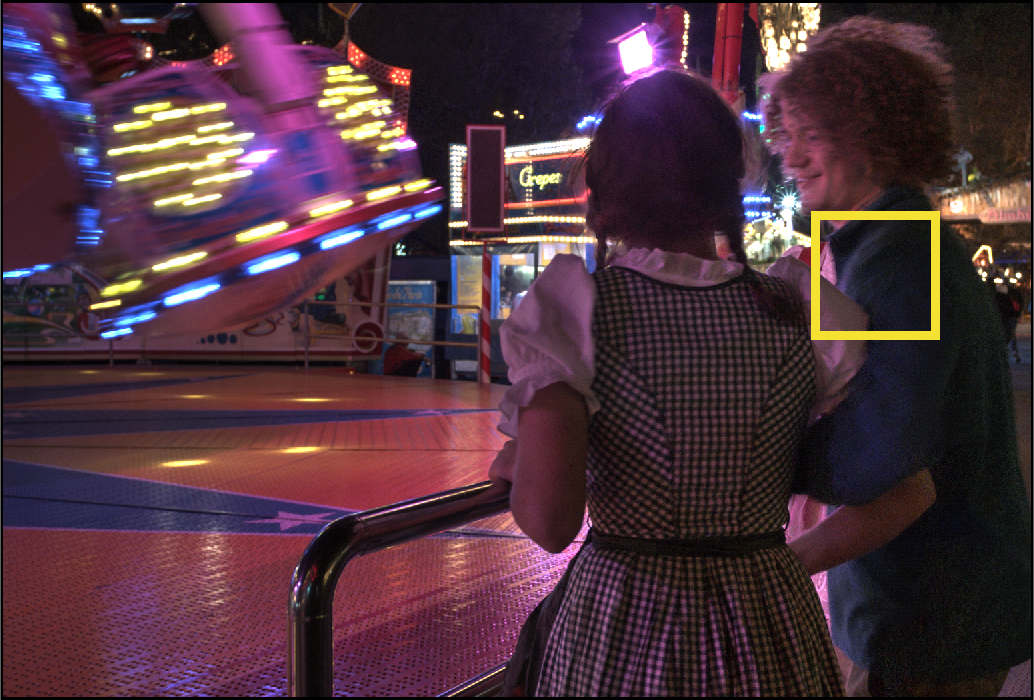}
			\end{subfigure}	
			\begin{subfigure}{\wteaser\textwidth}
				\includegraphics[height=\hteaser]{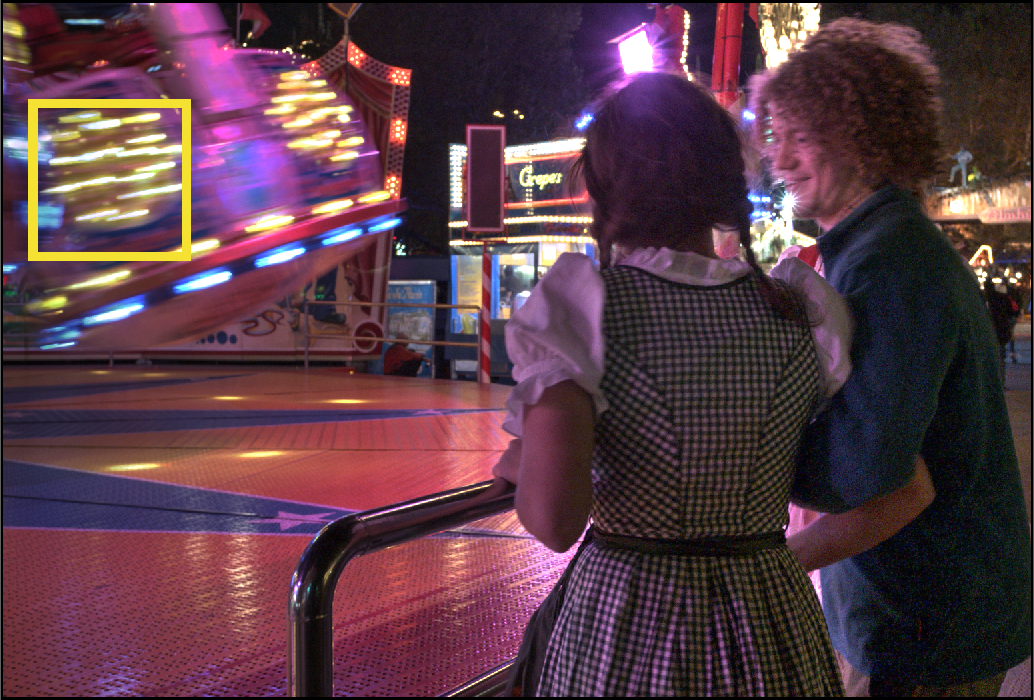}
			\end{subfigure}	
			\begin{subfigure}{\wteaser\textwidth}
				\includegraphics[height=\hteaser]{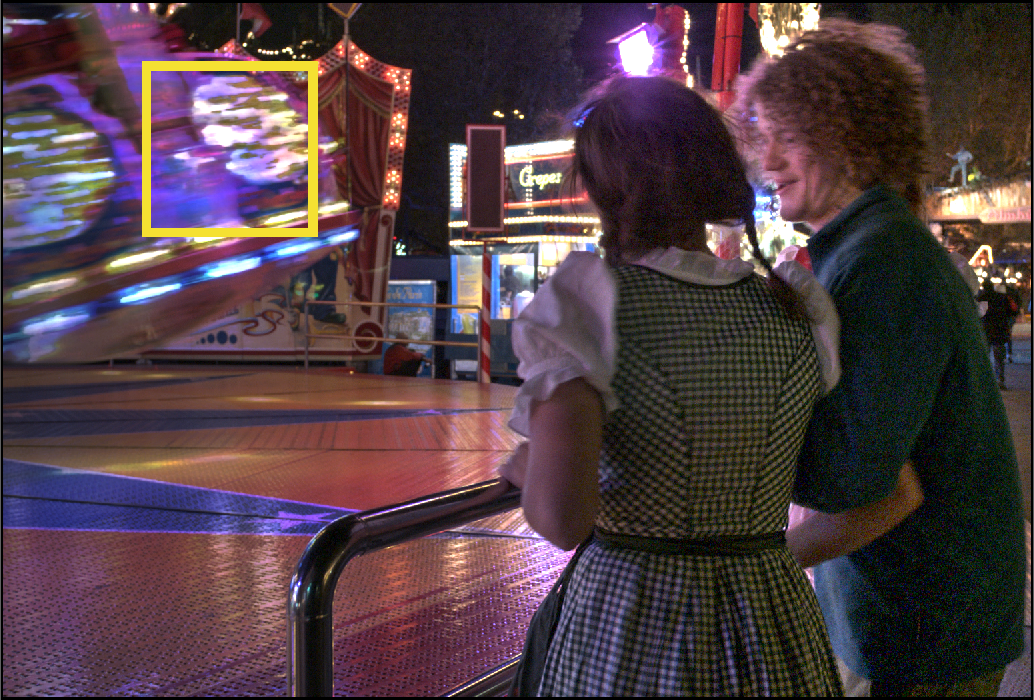}
			\end{subfigure}	
			\newline
			%
			\begin{subfigure}{\wteasers\textwidth}
				\includegraphics[height=\hteasers]{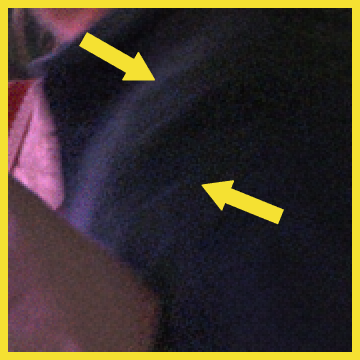}\vskip \mcaption
				\caption*{\: Chen \cite{chen2021hdr}} 
			\end{subfigure}	
			\begin{subfigure}{\wteasers\textwidth}
				\includegraphics[height=\hteasers]{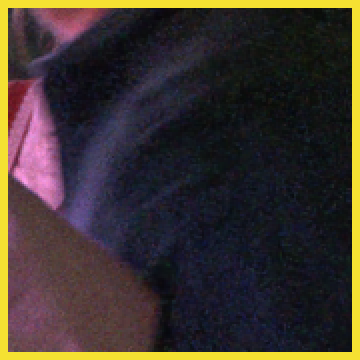}\vskip \mcaption
				\caption*{Ours} 
			\end{subfigure}	
			\begin{subfigure}{\wteasers\textwidth}
				\includegraphics[height=\hteasers]{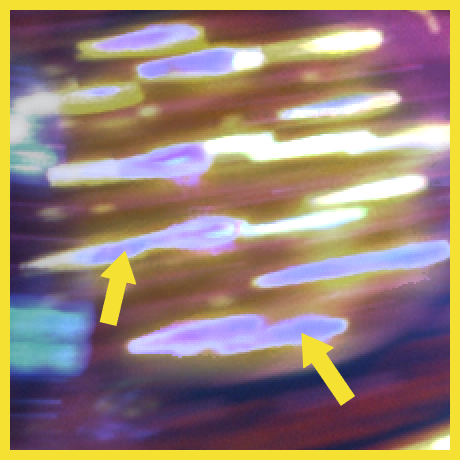}\vskip \mcaption
				\caption*{\: Chen \cite{chen2021hdr}} 
			\end{subfigure}	
			\begin{subfigure}{\wteasers\textwidth}
				\includegraphics[height=\hteasers]{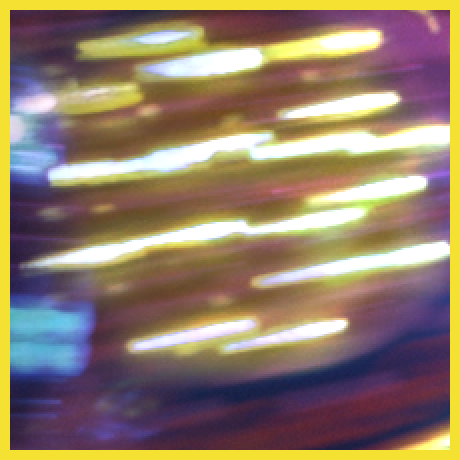}\vskip \mcaption
				\caption*{Ours} 
			\end{subfigure}	
			\begin{subfigure}{\wteasers\textwidth}
				\includegraphics[height=\hteasers]{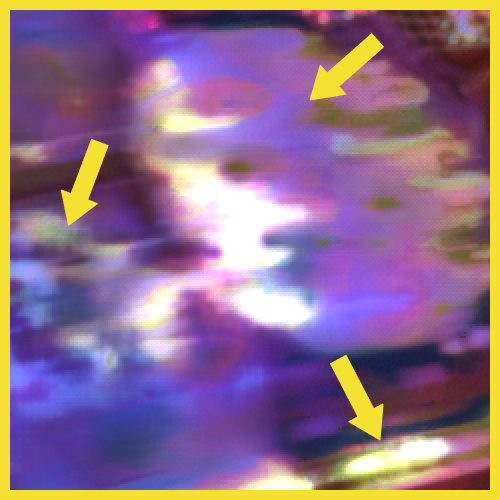}\vskip \mcaption
				\caption*{\: Chen \cite{chen2021hdr}} 
			\end{subfigure}	
			\begin{subfigure}{\wteasers\textwidth}
				\includegraphics[height=\hteasers]{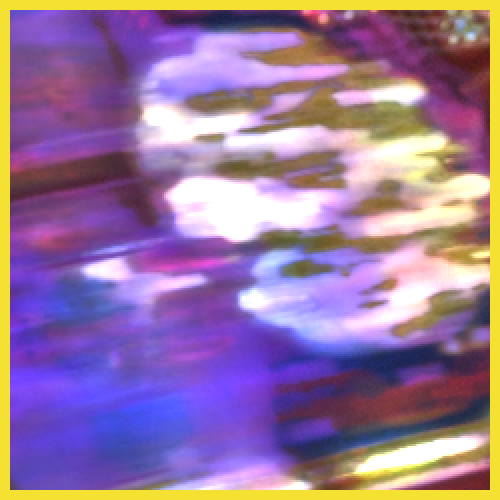}\vskip \mcaption
				\caption*{Ours} 
				
			\end{subfigure}	
		\end{minipage}
		\caption{Visual comparisons of the proposed and Chen \etal's method \cite{chen2021hdr}. Our framework generates more detailed texture in regions with saturation and motions by utilizing luminance information.}\vskip -2.6mm
		\label{fig:teaser}
	\end{figure}

	\section{Introduction}
	\label{sec:intro}
	
	As diverse videos become easily accessible through video-on-demand services, demands for high-quality video content with high resolution and high dynamic range (HDR) are naturally increasing. HDR content can provide a rich viewing experience by displaying high contrast and a broad range of colors. While HDR displays are already ubiquitous, there is still a lack of HDR content available for delivery.
	
	Whereas HDR imaging techniques for still images have been actively studied \cite{debevec1997recovering, an2014probabilistic, oh2015robust, kalantari2017deep, wu2018deep, yan2019attention, prabhakar2020towards, banterle2006inverse, rempel2007ldr2hdr, marnerides2018expandnet, endoSA2017, eilertsen2017hdr, alghamdi2019reconfigurable, liu2020single, Yang_2018_CVPR}, those for videos have been relatively overlooked. Some early methods \cite{unger2007high, tocci2011versatile, kronander2014unified} proposed well-designed hardware systems for direct HDR video acquisition, but these systems mostly require a sophisticated design and high cost. Since Kang \etal~\cite{kang2003high} reconstructed an HDR video using a low dynamic range (LDR) image sequence with alternating exposures, this approach has been commonly adopted. Specifically, an HDR frame is generated from a corresponding LDR frame and its neighboring frames, whose exposures alternate between two or three values, after motions between the frames are compensated. For instance, Mangiat \etal \cite{mangiat2010high, mangiat2011spatially} performed block-based motion estimation and then refined the results using color similarity and filtering, respectively. Kalantari \etal~\cite{kalantari2013patch} synthesized multi-exposure images at each time step using patch-based optimization and merged them into an HDR frame. These approaches mostly require considerable time for the optimization process and suffer from artifacts resulting from inaccurate alignment. 
	
	With the development of deep learning, convolutional neural network (CNN)-based HDR video reconstruction methods have been proposed. Similar to the previous HDR imaging method~\cite{kalantari2017deep}, Kalantari \etal~\cite{kalantari2019deep} aligned neighboring frames to the middle frame using optical flow and combined them with a simple merging network. More recently, Chen \etal~\cite{chen2021hdr} trained a two-step alignment network using optical flow and deformable convolution with large datasets. These methods significantly improved the quality of resulting HDR videos, but they still show some distortions and ghosting artifacts caused by optical flow estimation error, especially when motion exists in saturated regions or motion is fast. (See Fig.~\ref{fig:teaser}.)
	
	In this paper, we propose a luminance-based alignment network (LAN) for HDR video composition. The LAN has a dual-path architecture, where an alignment module registers motions between frames using attention \cite{vaswani2017attention}, and a hallucination module generates fine details, especially for saturated regions. In the alignment module, we rearrange a neighboring frame based on the attention score between itself and the reference frame. However, by applying naive attention, the system may find similar pixels solely by color. 
	For content-based matching between two frames, we extract key and query features using only the luminance channel, which contains essential information on edge and structure. Here, downsampled frames are used as input to alleviate memory consumption. However, alignment cannot be perfect because downsampled inputs lack high-frequency information, and saturation hinders precise matching. Thus, the hallucination module fills in the missing details, especially in saturated areas using full-sized inputs, where we use gated convolution \cite{yu2019free} with a mask that represents the image brightness. This mask enables the adaptive operation for very dark regions as well as highlighted areas. The features from each module are fused in an adaptive blending layer, which determines the contribution of each feature in terms of spatial dimension. Finally, we adopt a temporal loss for the resulting video to have consistent motions. Extensive experiments demonstrate that our method produces a clean HDR video from an LDR video with alternating exposures.
	
	The main contributions of our paper are summarized as follows:	
	\begin{itemize}[leftmargin=1.6em]
		\setlength\itemsep{0.5pt}
		\item  We introduce an end-to-end HDR video reconstruction framework that performs a precise motion alignment using the proposed luminance-based alignment network (LAN). The LAN consists of two novel components: an alignment module and a hallucination module.
		\item The proposed alignment module performs content-based alignment by rearranging a neighboring frame feature (\textit{value}) according to the attention score between content features from the neighboring frame (\textit{key}) and the reference frame (\textit{query}).
		\item The proposed hallucination module generates sharp details using adaptive convolution, especially for very bright or dark regions.
		\item We present a temporal loss to produce temporally coherent HDR videos without flickering.
	\end{itemize}
	\begin{figure*}[t!]
		\begin{center}
			\includegraphics[width=\linewidth]{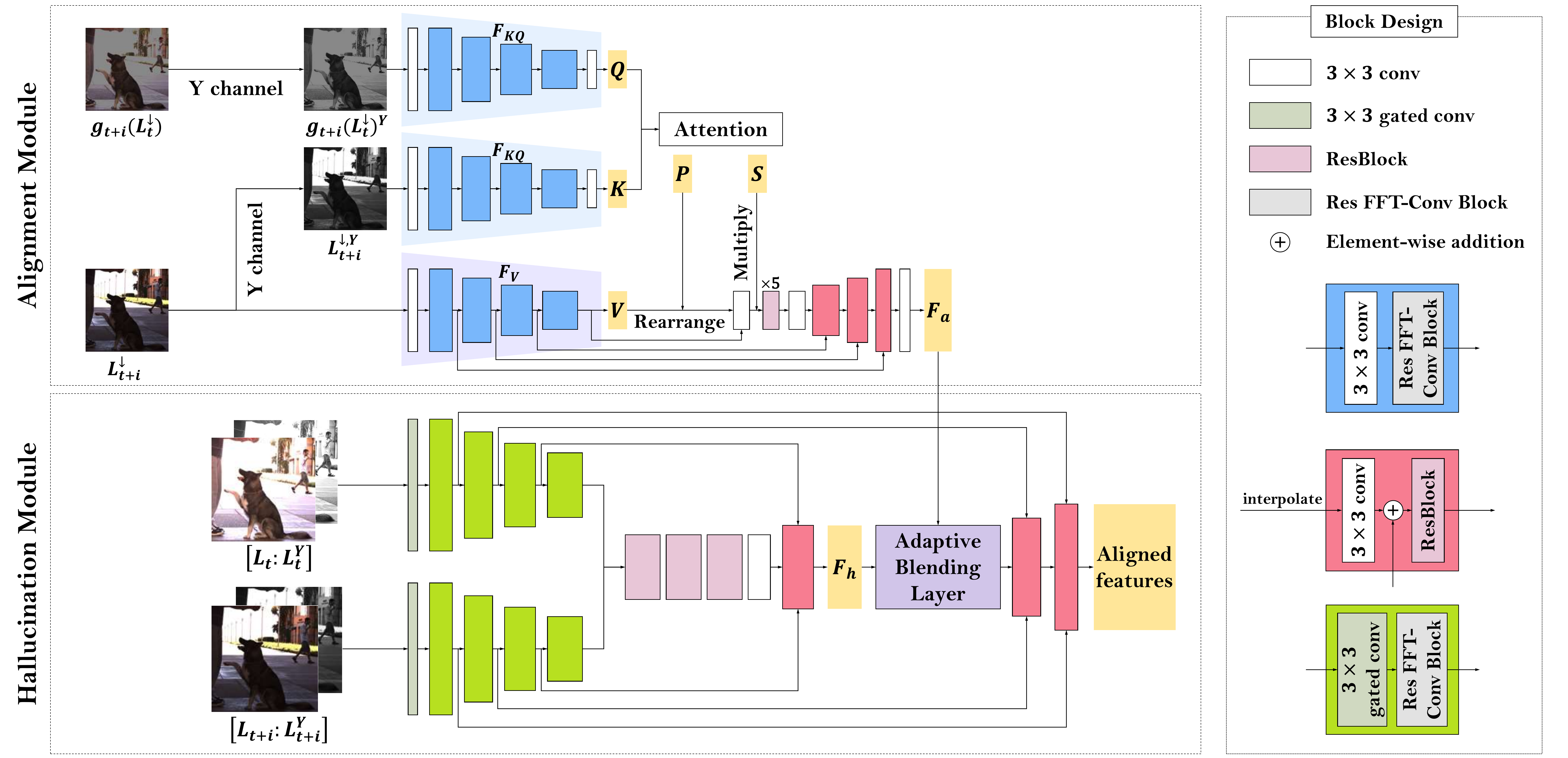}
		\end{center}
		\caption{Overview of the proposed luminance-based alignment network (LAN).} 
		\label{fig:overview}
	\end{figure*}
	\section{Related Work} \label{sec:relatedwork}
	\subsection{Multi-exposure HDR image reconstruction}
	HDR video reconstruction is similar to multi-exposure HDR imaging, a task of producing an HDR image from multiple differently exposed LDR inputs, in that they both aim to merge multi-exposure LDR frames with motions into a clean HDR frame. Most HDR imaging approaches focus on aligning input images to avoid ghosting artifacts. Some methods \cite{khan2006ghost, an2014probabilistic, heo2010ghost, grosch2006fast, zhang2012gradient, oh2015robust, lee2014ghost} assumed globally registered inputs and tried to detect and reject shifted pixels. Instead of simply discarding moving parts, several works performed more sophisticated motion compensation using optical flow \cite{kang2003high, zimmer2011freehand} and patch-based matching \cite{sen2012robust, hu2013hdr}. However, most of these methods are slow due to the optimization process and often fail to handle large motions.
	
	With the advancement of deep learning, HDR imaging networks have also been proposed. Kalantari \etal~\cite{kalantari2017deep} and Wu \etal~\cite{wu2018deep} proposed methods that align LDRs by flow-based motion compensation and homography transformation, respectively, and then merge them by simple CNNs. Pu \etal~\cite{pu2020robust} presented a deformable convolution-based pyramidal alignment network, while Prabhakar \etal~\cite{prabhakar2021labeled} trained a fusion network using both labeled and unlabeled data. Yan \etal~\cite{yan2019attention, yan2020deep} aligned input images using spatial attention and non-local network, respectively. Note that these approaches differ significantly from ours in that they implicitly induce their attention maps to select valid areas through training. On the other hand, we compute the correlation between the key and query frames and then reconstruct aligned results based on the estimated scores.
	It is possible to apply the aforementioned methods to HDR video composition frame by frame, but temporal consistency is hardly guaranteed.
	
	\subsection{HDR video reconstruction}
	Instead of directly acquiring HDR videos using specialized camera systems \cite{unger2007high, tocci2011versatile, kronander2014unified}, which are very expensive approaches, many methods exploit LDR videos consisting of frames with spatially or temporally varying exposures. However, most approaches using dual- or tri-exposure images \cite{cogalan2022learning, jiang2021hdr} experience spatial artifacts. Meanwhile, similar to the multi-exposure HDR image fusion, most of the methods using a sequence with alternating exposures align multiple frames and combine them to produce an HDR frame. Kang \etal~\cite{kang2003high} proposed the first algorithm in this category, which synthesizes aligned multi-exposure images at each time step using global and local registration and merges them into an HDR frame.  Kalantari \etal~\cite{kalantari2013patch} also reconstructed missing images with different exposures based on patch-based optimization~\cite{sen2012robust}. Mangiat \etal~\cite{mangiat2010high} adopted block-based motion estimation and additionally refined motion vector, which was improved by Mangiat \etal~\cite{mangiat2011spatially} with the addition of HDR filtering for block artifact removal. Meanwhile, Gryaditskaya \etal~\cite{gryaditskaya2015motion} proposed an adaptive metering algorithm to minimize motion artifacts. Li \etal~\cite{li2016maximum} avoided correspondence estimation by formulating HDR video reconstruction as a maximum a posteriori (MAP) estimation problem, where foreground and background are synthesized separately and then combined together. These methods are mostly time-consuming and often display visible artifacts.
	
	The recent work of Kalantari \etal~\cite{kalantari2019deep} first presented a CNN-based framework that consists of a flow network and a merging network, similar to the HDR imaging method in \cite{kalantari2017deep}. However, the flow estimation error often leads to undesirable artifacts in the resulting images. Chen \etal~\cite{chen2021hdr} trained a coarse-to-fine network, including a pyramid, cascading, and deformable alignment module~\cite{wang2019edvr} with newly collected datasets. This work requires global alignment as preprocessing and still uses optical flow for coarse alignment. Despite the introduction of various alignment strategies, these approaches often suffer from artifacts from misalignment in saturated areas and temporal inconsistency due to frame-by-frame reconstruction. In contrast, we perform luminance-based alignment without pre-alignment and improve temporal coherency by adding a temporal loss.
	
	\section{Proposed Method}
	Given an LDR video with alternating exposures, our goal is to generate a high-quality HDR video. In order to produce an HDR frame $H_t$, we use $2N+1$ consecutive LDR frames $\{L_{t-N}, \cdots, L_t, \cdots, L_{t+N}\}$, which consist of a corresponding LDR frame and its neighbors. Following the previous work \cite{chen2021hdr}, five frames are used as input in the case of having two alternating exposures ($N=2$), and seven frames for the three ($N=3$). For brevity, this paper mostly addresses the case of two alternating exposures and discusses the extension to the three in Section \ref{extension}.
	
	Before feeding the inputs into our framework, we map the LDR frames to the HDR using gamma correction with $\gamma=2.2$: 
	\begin{equation}
		X_t = {L_t^\gamma}/{e_t},
	\end{equation} 
	where $e_t$ is the exposure time of $L_t$. We then construct a $6$-channel input by concatenating $L_t$ and $X_t$ along the channel dimension. In the rest of the paper, we denote the LDR input as $L_t$ for brevity, though its gamma-transformed HDR image $X_t$ is actually used together as the input.  
	
	Our framework is composed of weight-sharing luminance-based alignment networks (LANs) and a merging network. Each LAN aligns the motion of the supporting frame $L_{t+i}$ to that of the reference $L_{t}$, and the merging network combines the features from the LANs into an HDR frame $H_{t}$. As shown in Fig.~\ref{fig:overview}, the LAN has two different modules. The alignment module calculates the attention score between the reference frame (\textit{query}) and the supporting frame (\textit{key}) and then aligns the supporting frame (\textit{value}) based on this score. Since the alignment module can produce incomplete content in saturated areas, the hallucination module generates fine details in the desired regions. The features from the two modules are fused in an adaptive blending layer. The outputs of the LANs are concatenated and fed to the merging network for HDR frame composition. The details of each network are described in the following sections.
	\subsection{Luminance-based Alignment Network (LAN)}
	\noindent \textbf{Alignment module} The core idea of the alignment module is to calculate attention between a neighboring frame (\textit{key}) and a reference frame (\textit{query}) based on their content and then rearrange the neighboring frame (\textit{value}) using the estimated attention. To alleviate memory usage and computation for attention operation, we input low-resolution images downsampled by a factor of $4$. Here, we first adjust the exposure of the downsampled reference frame ${L}_{t}^{\downarrow}$ to that of the downsampled neighboring frame ${L}_{t+i}^{\downarrow}$: $g_{t+i}({L}_{t}^{\downarrow}) = clip\left({L}_{t}^{\downarrow}\left({{e_{t+i}}/{e_t}}\right)^{{1}/{\gamma}}\right)$,
	where $e_t$ and $e_{t+i}$ denote the exposure time of ${L}_{t}^{\downarrow}$ and ${L}_{t+i}^{\downarrow}$, respectively. Then, after conversion to YCbCr space, only Y channels of each image, ${L}_{t+i}^{\downarrow , Y}$ and $g_{t+i}({L}_{t}^{\downarrow})^Y$, are fed to the key/query extractor $F_{KQ}$. The $F_{KQ}$ can extract key and query features focusing on content rather than superficial color information.
	
	The key and query features are then unfolded into patches to reflect neighbors. Likewise, the downsampled neighboring frame ${L}_{t+i}^{\downarrow}$ is embedded through a value extractor $F_{V}$ and unfolded to construct value features. The key $K$, query $Q$, and value $V$ for attention operation can be represented as a set of patches: $K=\{\bm{k}_i,i\in[1,n]\}$, $Q=\{\bm{q}_i,i\in[1,n]\}$, and $V=\{\bm{v}_i,i\in[1,n]\}$, respectively. Each patch has the dimension of $3\times 3 \times C$, and $n$ is the number of pixels in the key/query/value features.
	
	To estimate correlation between $Q$ and $K$, patch-wise cosine similarity is calculated as:
	\begin{equation}
		c_{i,j}=\langle \frac{\bm{q}_i}{\norm{\bm{q}_i}},\frac{\bm{k}_j}{\norm{\bm{k}_j}} \rangle.
	\end{equation}
	In order to prevent blurry results, we exploit the most correlated (top-$1$) element, constructing an index map $I=\{{p}_i,i\in[1,n]\}$ and a confidence map $S=\{{s}_i,i\in[1,n]\}$. For each query patch $\bm{q}_i$, an index of the most correlated key patch ${p}_i$ and its confidence score ${s}_i$ are determined as:
	\begin{equation}
		{p}_i = \argmax _j c_{i,j},
	\end{equation}
	\begin{equation}
		{s}_i = \max_j c_{i,j}.
	\end{equation}
	Then, rearranged value features $V'$ are obtained, where $\bm{v}_{{p}_i}$ is located at the $i$-th location. To prevent information loss in the case of mismatching, we concatenate the original value features $V$ with the rearranged features $V'$ and then multiply the confidence map $S$. The adjusted features are upsampled to the resolution of the input ${L}_{t+i}^{\downarrow}$ progressively through upsampling blocks, resulting in the aligned features $F_a$.  
	
	\newcommand{\wmap}{0.156}
	\newcommand{\hmap}{1.53cm}
	
	\begin{figure}[t]
		\centering
		\begin{subfigure}{\wmap\textwidth}
			\includegraphics[height=\hmap]{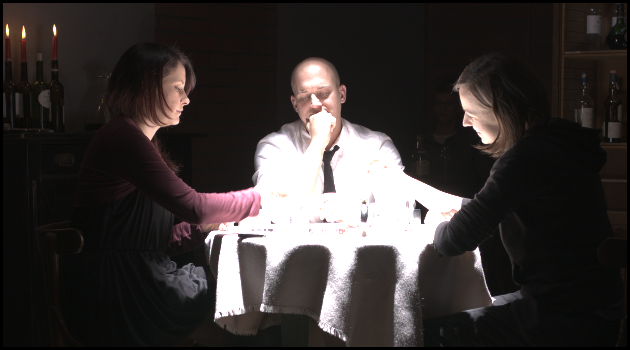} \vskip \mcaption
			\caption*{$L_{t}$} 
		\end{subfigure}	
		\begin{subfigure}{\wmap\textwidth}
			\includegraphics[height=\hmap]{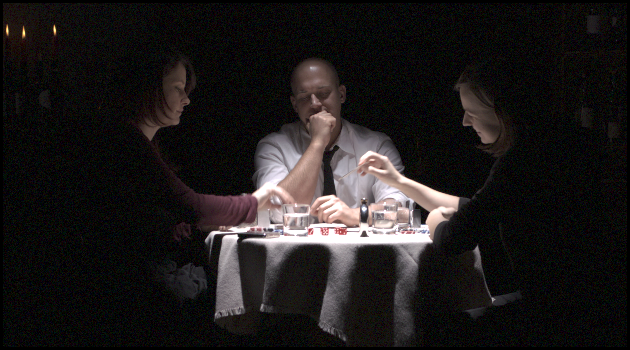} \vskip \mcaption
			\caption*{$L_{t+1}$} 
		\end{subfigure}	
		\begin{subfigure}{\wmap\textwidth}
			\includegraphics[height=\hmap]{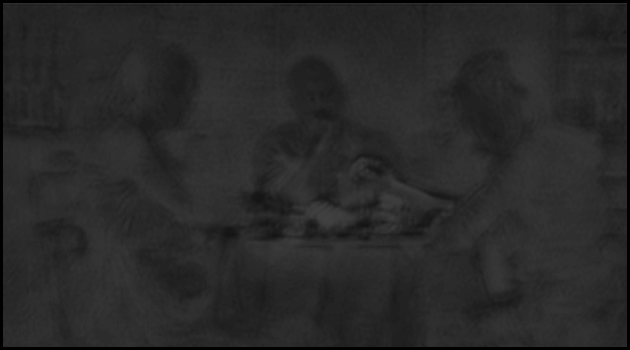} \vskip \mcaption
			\caption*{$M$}
		\end{subfigure}	
		
		\caption{An example of a blending map $M$ when aligning a neighboring frame $L_{t+1}$ to the reference $L_t$ with saturation. The darker the map, the more features from the hallucination module $F_h$ contribute to reconstruction.}
		\label{fig:map}
	\end{figure}
	
	\noindent \textbf{Hallucination module} It is difficult to perform the alignment when the reference image has severely saturated areas such as the one in Fig.~\ref{fig:map}. Also, downsampled inputs lack sharp details while improving efficiency. The hallucination module aims to fill in the missing content resulting from saturation and generate high-frequency details. To this end, we utilize gated convolution \cite{yu2019free}, which is widely adopted in inpainting networks, to restore specific saturated regions. Also, masks indicating saturated parts as well as the original images are used as input to an encoder-decoder architecture. According to the brightness information of the input masks, the gated convolution performs adaptively with respect to spatial and channel dimensions. Given input features $F_i$, the output of the gated convolution $F_{o}$ is formulated as: 
	\begin{equation} 
		F_{o} = \rho(\sum\sum W_f \cdot F_i)\odot\sigma(\sum\sum W_g \cdot F_i),
	\end{equation}
	where $\odot$ denotes element-wise multiplication. $\rho$ and $\sigma$ are the exponential linear unit and sigmoid, respectively. $W_f$ denotes a convolutional filter for the feature, and $W_g$ is the one for gating. Thus, gated convolution enables dynamic operation that is aware of saturation.
	We use the Y channel of each image as a mask to provide explicit but continuous information about luminance. The luminance channel indicates dark regions as well as saturated parts, thus helping to suppress noise in dark areas. Analysis of the input to the hallucination module is presented in Section~\ref{ablations}.
	
	\noindent \textbf{Adaptive blending layer} Aligned features from the alignment module $F_a$ are fused with intermediate features of the hallucination module $F_h$ through an adaptive blending layer. For the two modules to complement each other, the adaptive blending layer estimates a blending map $M\in \mathbb{R}^{H\times W \times 1}$ in the range $[0,1]$, where $H\times W$ is the spatial resolution of $F_a$ and $F_h$. The output features of the fusion $ F_{out} $ are represented as follows:
	\begin{equation} 
		F_{out} = (1-M) \odot F_h + M \odot (F_h+F_a).
	\end{equation}
	When the alignment is challenging in saturated parts due to missing information, the contribution of $F_a$ becomes small, represented as a low value in the map $M$. (See an example in Fig.~\ref{fig:map}.) The $ F_{out} $ passes through the remaining upsampling blocks to recover its spatial size.
	
	\subsection{Merging Network}
	Since the proposed LANs output well-aligned features, our merging network has a simple structure. The aligned features are concatenated along the channel axis and fed to a convolutional layer and five Res FFT-Conv Blocks \cite{mao2021deep}. This network can cover the entire frame globally by performing convolutions in the frequency domain. After the final convolutional layer with a sigmoid activation function, the final HDR frame $H_t$ is constructed.
	
	\subsection{Training Loss}
	Since HDR images are usually displayed after tonemapping, we calculate loss functions between the tonemapped predicted HDR frame $\mathcal{T}(\hat{H_t})$ and the tonemapped ground truth HDR image $\mathcal{T}(H_t)$ using the differentiable $\mu$-law: 
	\begin{equation} 
		\mathcal{T}(H) = \frac{\log(1+\mu H)}{\log(1+H)},
	\end{equation}
	where $\mu$ is set to $ 5000 $.
	
	\noindent \textbf{$\mathcal{L}_1$ loss} We adopt $\mathcal{L}_1$ loss which is defined as: 
	$	\mathcal{L}_{1}=\norm{\mathcal{T}(\hat{H_t}) - \mathcal{T}(H_t)}_1 $.

	\noindent \textbf{Perceptual loss} Using the feature map from the pre-trained VGG-19 network $\phi(\cdot)$, we calculate perceptual loss $\mathcal{L}_{per}$ as: 
	$	\mathcal{L}_{per}=\norm{\phi(\mathcal{T}(\hat{H_t})) - \phi(\mathcal{T}(H_t))}_2^2$.
	
	\noindent \textbf{Frequency loss} We also utilize a frequency loss which has been shown to be effective in recent low-level vision tasks \cite{fuoli2021fourier, cho2021rethinking, wang2022artifact}. The frequency loss $L_{freq}$ is defined as:
	$	\mathcal{L}_{freq}=\norm{\mathcal{F}(\mathcal{T}(\hat{H_t})) - \mathcal{F}(\mathcal{T}(H_t))}_1$,
	where $\mathcal{F}$ is the fast Fourier transform (FFT).
	
	\noindent \textbf{Temporal loss} To generate a perceptually natural video, temporal consistency without abrupt change is crucial. Since every HDR frame is constructed independently, we present our temporal loss $\mathcal{L}_{temp}$ to improve temporal consistency. Specifically, we enforce a difference map between the consecutive tonemapped outputs to be similar to the ground truth using the Charbonnier penalty function~\cite{charbonnier1994two}. The temporal loss for this enforcement is defined as: 
	\begin{equation} 
		\footnotesize \mathcal{L}_{temp}=\sqrt{ \norm{(\mathcal{T}(\hat{H_t})-\mathcal{T}(\hat{H}_{t-1})) - (\mathcal{T}(H_t)-\mathcal{T}(H_{t-1}))}_2^2 +\epsilon^2 },
	\end{equation} 
	where $\epsilon$ is set as $10^{-3}$. 
	
	Our total loss is represented as: $\mathcal{L}_{total}=\lambda_{1}\mathcal{L}_{1} + \lambda_{per}\mathcal{L}_{per} + \lambda_{freq}\mathcal{L}_{freq} + \lambda_{temp}\mathcal{L}_{temp}$, where $\lambda_{1}=1$, $\lambda_{per}=0.1$, $\lambda_{freq}=0.1$, and $\lambda_{temp}=0.1$.
	
	\subsection{Extension to Three Exposures} \label{extension}
	Given an LDR sequence with alternating exposures in three levels, seven LDR frames are used to compose an HDR frame. Our LAN aligns each pair of a supporting frame and the reference, and the merging network combines them in the same manner.
	
	\begin{table*}[t!]
		\caption{Quantitative comparisons of our method with other state-of-the-arts on the Cinematic Video dataset \cite{froehlich2014creating}.}
		\label{tab:synthetic}
		\centering 
		\resizebox{\textwidth}{!}{%
			\begin{tabular}{l|ccccc|ccccc}
				\hline 
				& \multicolumn{5}{c|}{2 Exposures}                                                                                                                                                                                                                                                             & \multicolumn{5}{c}{3 Exposures}                                                                                                                                                                                                                                                            \\ \cline{2-11} 
				Methods   & \multicolumn{1}{l}{PSNR\textsubscript{$T$}} & \multicolumn{1}{l}{SSIM\textsubscript{$T$}} & \multicolumn{1}{l}{PSNR\textsubscript{$PU$}} & \multicolumn{1}{l}{SSIM\textsubscript{$PU$}} & \multicolumn{1}{l|}{HDR-VDP-2} & \multicolumn{1}{l}{PSNR\textsubscript{$T$}} & \multicolumn{1}{l}{SSIM\textsubscript{$T$}} & \multicolumn{1}{l}{PSNR\textsubscript{$PU$}} & \multicolumn{1}{l}{SSIM\textsubscript{$PU$}} & \multicolumn{1}{l}{HDR-VDP-2} \\ \hline
				Kalantari \cite{kalantari2013patch}    & \blue{\textbf{37.51}}   & 0.9016 & \blue{\textbf{39.13}}    & \red{\textbf{0.9319}} & 60.16   & 30.36   & 0.8133 & 32.53    & 0.8251   & 57.68 \\
				Kalantari \cite{kalantari2019deep}  & 37.06   & \blue{\textbf{0.9053}} & 39.10    & \blue{\textbf{0.9286}}    & \blue{\textbf{70.82}}  & 33.21   & 0.8402  & 35.38    & \blue{\textbf{0.8582}}   & 62.44 \\
				Yan \cite{yan2019attention} & 31.65   & 0.8757  & 32.99    & 0.7523   & 69.05  & \blue{\textbf{34.22}}   & 0.8604 & 36.04    & 0.8357    & \blue{\textbf{66.18}} \\
				Prabhakar \cite{prabhakar2021labeled}   & 34.72   & 0.8761  & 36.06    & 0.8257 & 68.82   & 34.02   & 0.8633 & 36.19    & 0.8590   & 65.00 \\
				Chen \cite{chen2021hdr}     & 35.65   & 0.8949  & 37.12    & 0.8156   & \red{\textbf{72.09}}   & 34.15   & \red{\textbf{0.8847}} & \blue{\textbf{36.23}}    & 0.8357   & \red{\textbf{66.81}} \\
				Ours    & \red{\textbf{38.22}}   & \red{\textbf{0.9100}} & \red{\textbf{40.04}}   & 0.9039 & 69.15    & \red{\textbf{35.07}}   & \blue{\textbf{0.8695}} &\red{\textbf{37.22}}    & \red{\textbf{0.8666}}   & 65.42 \\ \hline
			\end{tabular}
		}
	\end{table*}

	\begin{table*}[t!]
		\caption{Quantitative comparisons of our method with other state-of-the-arts on the DeepHDRVideo dataset \cite{chen2021hdr}.}
		\label{tab:real}
		\centering 
		\resizebox{\textwidth}{!}{%
			\begin{tabular}{l|ccccc|ccccc}
				\hline 
				& \multicolumn{5}{c|}{2 Exposures}                                                                                                                                                                                                                                                             & \multicolumn{5}{c}{3 Exposures}                                                                                                                                                                                                                                                            \\ \cline{2-11} 
				Methods   & \multicolumn{1}{l}{PSNR\textsubscript{$T$}} & \multicolumn{1}{l}{SSIM\textsubscript{$T$}} & \multicolumn{1}{l}{PSNR\textsubscript{$PU$}} & \multicolumn{1}{l}{SSIM\textsubscript{$PU$}} & \multicolumn{1}{l|}{HDR-VDP-2} & \multicolumn{1}{l}{PSNR\textsubscript{$T$}} & \multicolumn{1}{l}{SSIM\textsubscript{$T$}} & \multicolumn{1}{l}{PSNR\textsubscript{$PU$}} & \multicolumn{1}{l}{SSIM\textsubscript{$PU$}} & \multicolumn{1}{l}{HDR-VDP-2} \\ \hline 
				Kalantari \cite{kalantari2013patch}  & 40.33   & 0.9409 & 43.71    & 0.9646   & 66.11    & 38.45   & 0.9489 & 42.35    & \blue{\textbf{0.9740}} & 57.31 \\
				Kalantari \cite{kalantari2019deep}   & 39.91   & 0.9329 & 43.31    & 0.9641  & 71.11    & 38.78   & 0.9331 & 41.80    & 0.9647  & 65.73 \\
				Yan \cite{yan2019attention}     & {40.54}   & 0.9452 &\blue{\textbf{45.33}}    & 0.9616 & 69.67 & \blue{\textbf{40.20}}   & \blue{\textbf{0.9531}}  & \blue{\textbf{42.36}}    & 0.9700    & \blue{\textbf{68.23}} \\
				Prabhakar \cite{prabhakar2021labeled}  & 40.21   & 0.9414 & 45.16    & 0.9593   & 70.27   & 39.48   & 0.9453  & 41.15    & 0.9666  & 65.93 \\
				Chen \cite{chen2021hdr}       & \red{\textbf{42.48}}   & \red{\textbf{0.9620}} & \red{\textbf{45.79}}    & \red{\textbf{0.9773}}  & \red{\textbf{74.80}}  & 39.44   & \red{\textbf{0.9569}}   & 41.57    & {0.9725} & 67.76 \\
				Ours & \blue{\textbf{41.59}}   & \blue{\textbf{0.9472}} & 44.43    & \blue{\textbf{0.9730}}   & \blue{\textbf{71.34}}   & \red{\textbf{40.48}}  & 0.9504 &\red{\textbf{42.38}}    & \red{\textbf{0.9755}}     & \red{\textbf{68.61}} \\ \hline
			\end{tabular}
		}
	\end{table*}

	\newcommand{\bigrowmargin}{-1.0mm}
	\newcommand{\rowmargin}{-1.7mm}
	
	\newcommand{\htemp}{3.45cm}
	\newcommand{\hseven}{1.24cm}
	\newcommand{\wseven}{0.0691}
	\newcommand{\wsevenlast}{0.051}
	\newcommand{\hmargin}{0.00mm}
	\begin{figure*}[t!]
		\begin{subfigure}{.0641\textwidth}
			\includegraphics[height=\htemp]{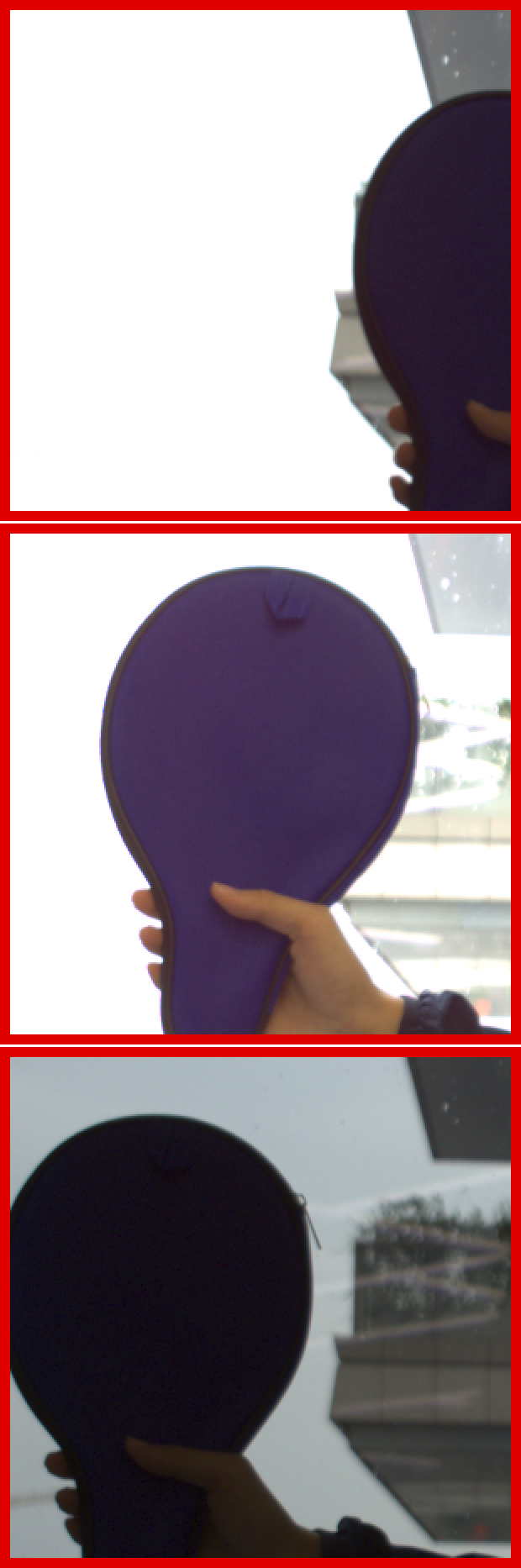} \vskip \mcaption \caption*{~\quad Input}
		\end{subfigure}
		\begin{subfigure}{.0641\textwidth}
			\includegraphics[height=\htemp]{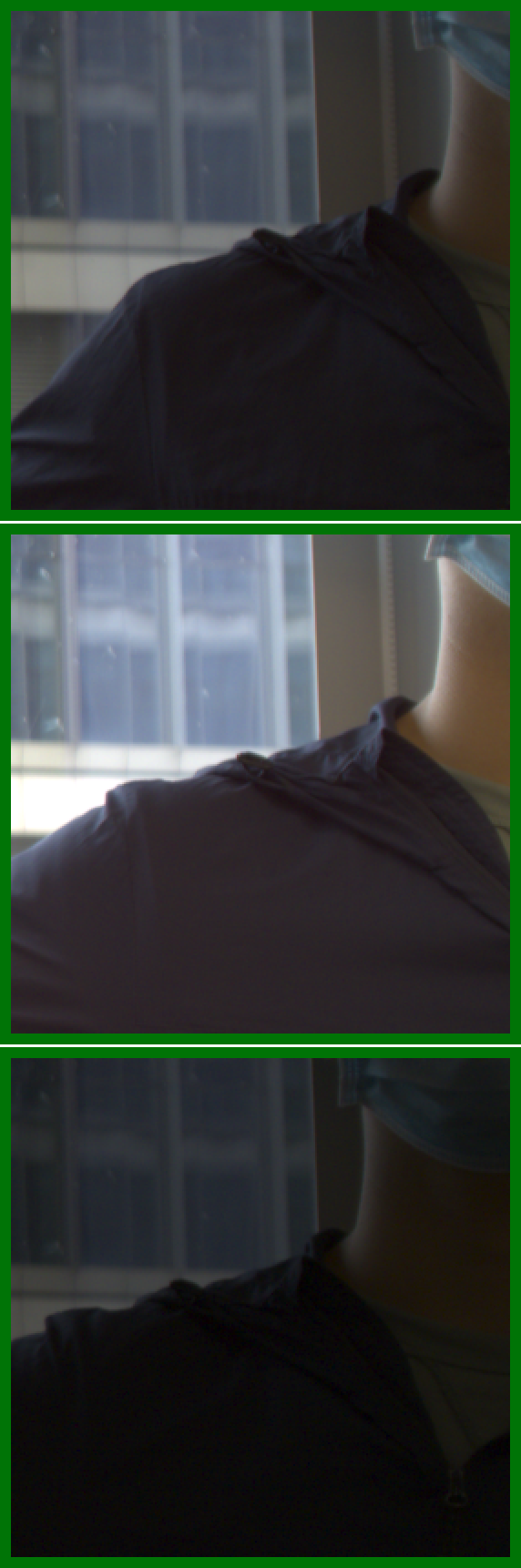} \vskip \mcaption  \caption*{Frames~~\quad\quad}
		\end{subfigure}
		\begin{subfigure}{.372\textwidth}
			\includegraphics[height=\htemp]{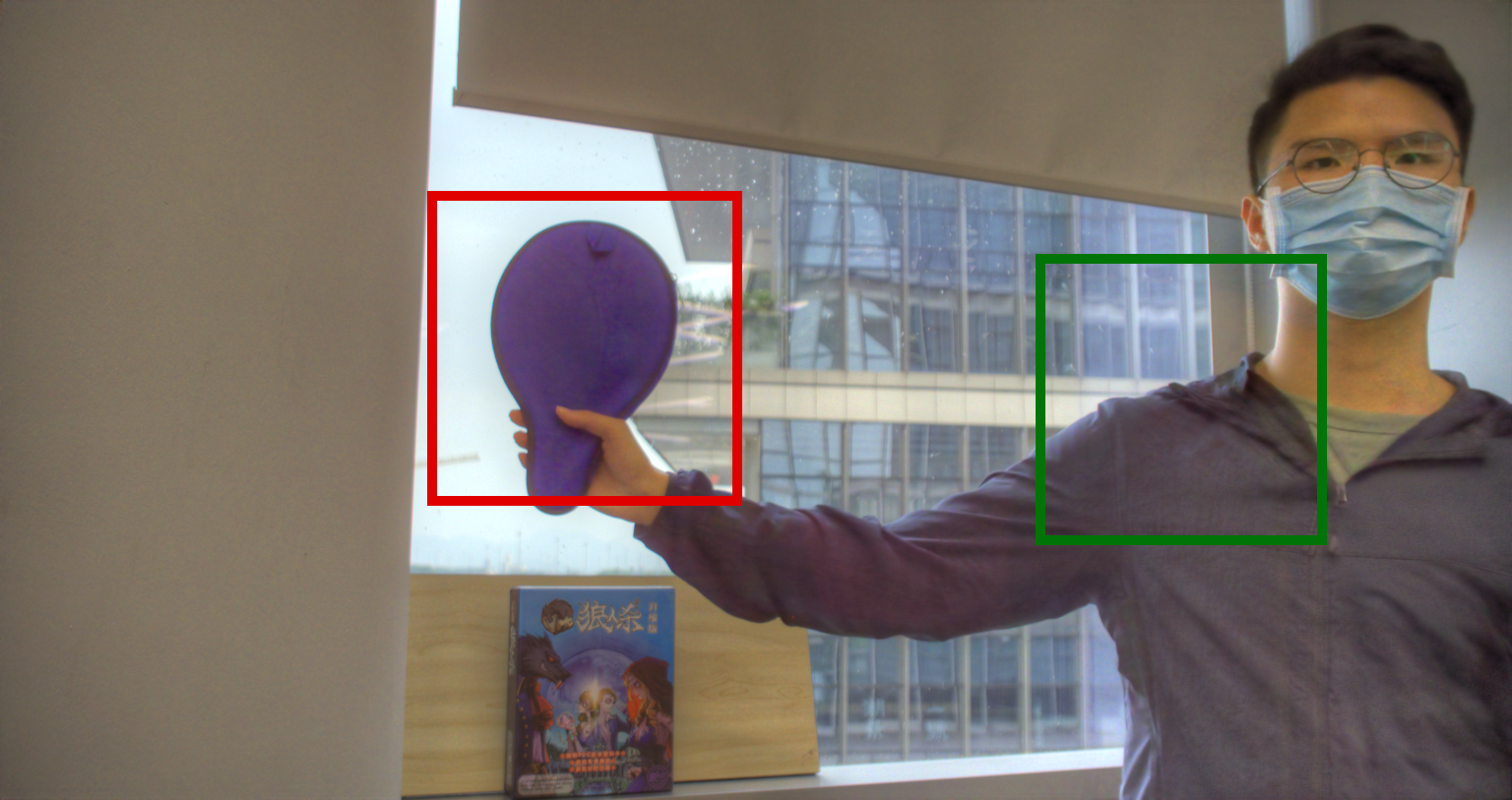} \vskip \mcaption  \caption*{Our Result}
		\end{subfigure}
		\begin{subfigure}{.0641\textwidth}
			\includegraphics[height=\htemp]{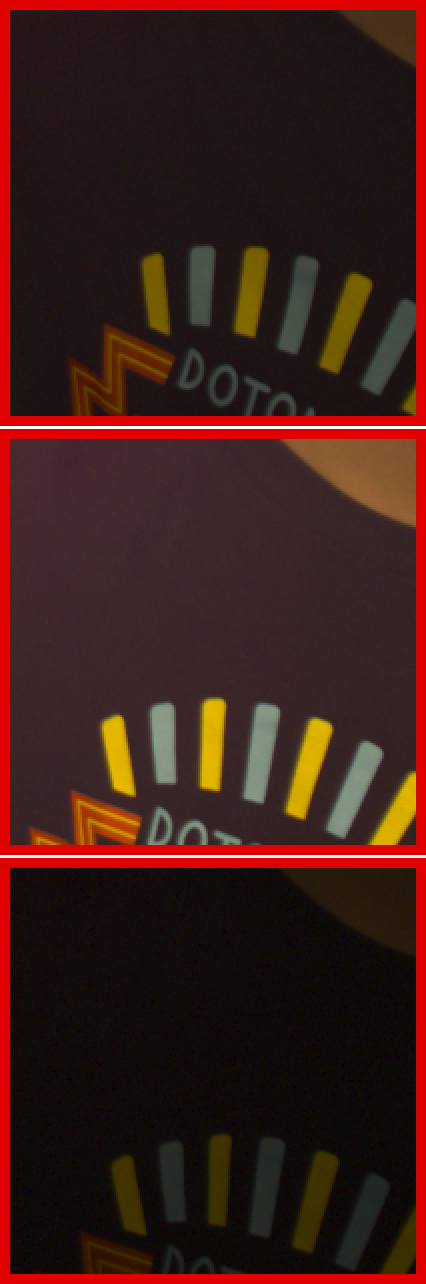} \vskip \mcaption  \caption*{~\quad Input}
		\end{subfigure}
		\begin{subfigure}{.0641\textwidth}
			\includegraphics[height=\htemp]{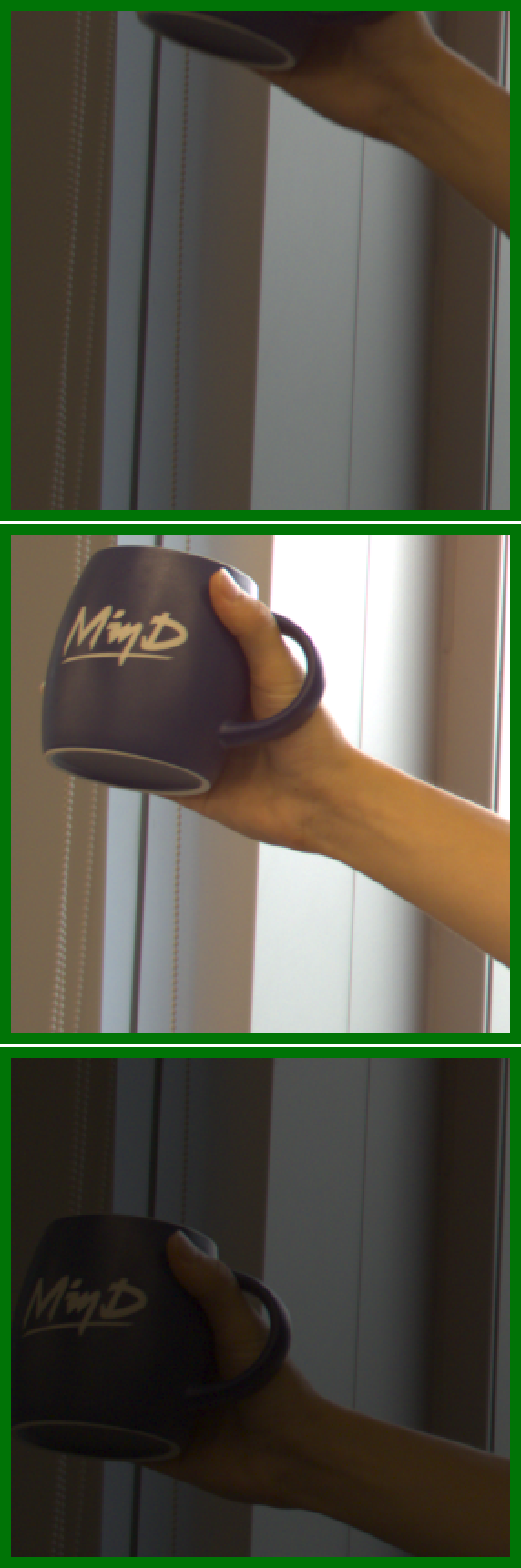} \vskip \mcaption  \caption*{Frames~~\quad\quad}
		\end{subfigure}
		\begin{subfigure}{.37\textwidth}
			\includegraphics[height=\htemp]{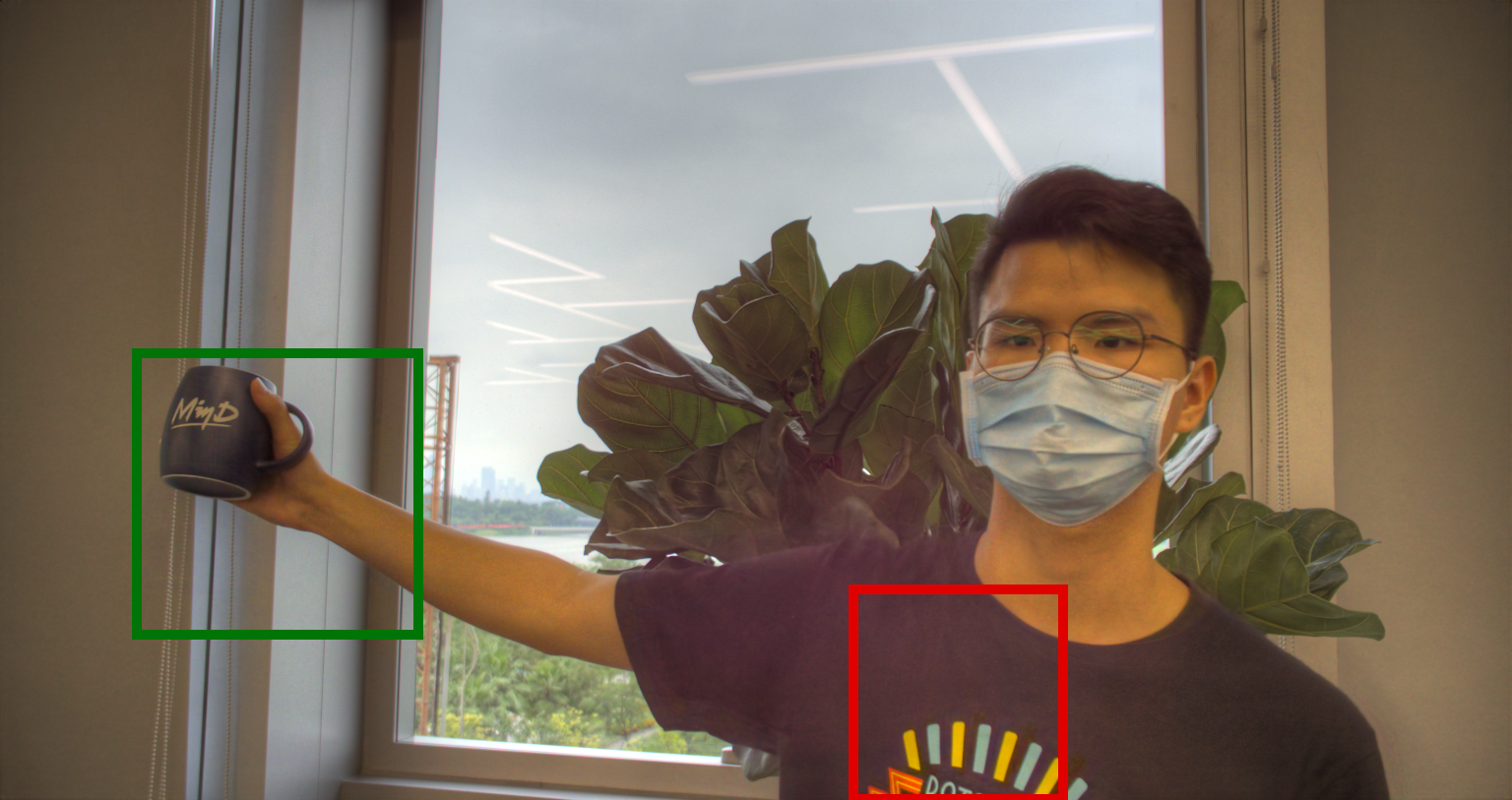} \vskip \mcaption  \caption*{Our Result}
		\end{subfigure}

		\begin{subfigure}{\wseven\textwidth}
			\includegraphics[height=\hseven]{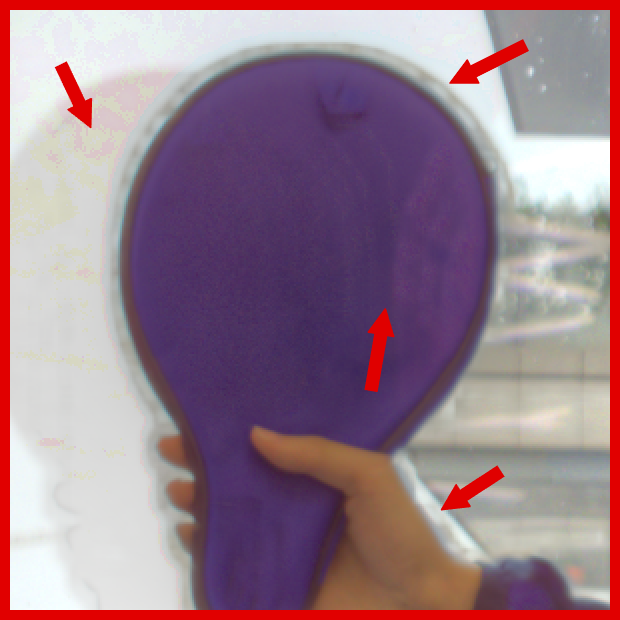}
		\end{subfigure}	
		\begin{subfigure}{\wseven\textwidth}
			\includegraphics[height=\hseven]{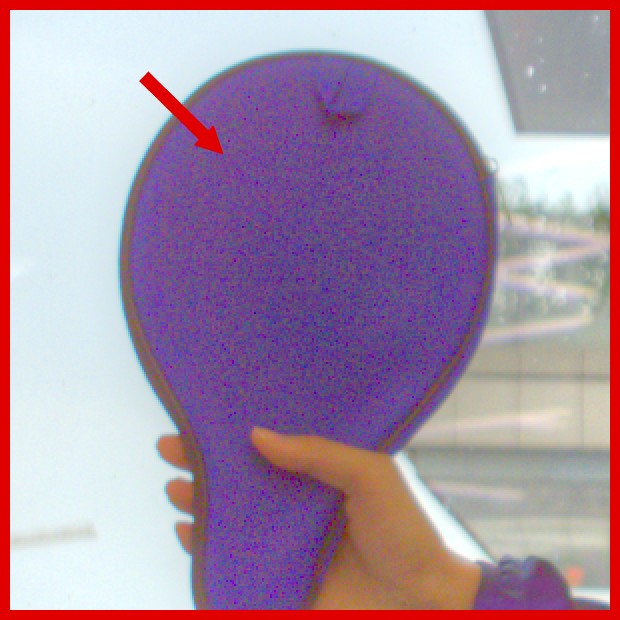}
		\end{subfigure}	
		\begin{subfigure}{\wseven\textwidth}
			\includegraphics[height=\hseven]{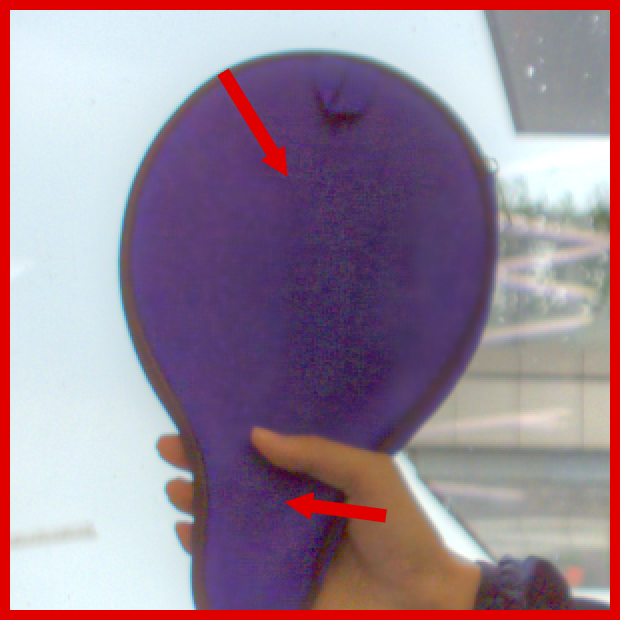}
		\end{subfigure}	
		\begin{subfigure}{\wseven\textwidth}
			\includegraphics[height=\hseven]{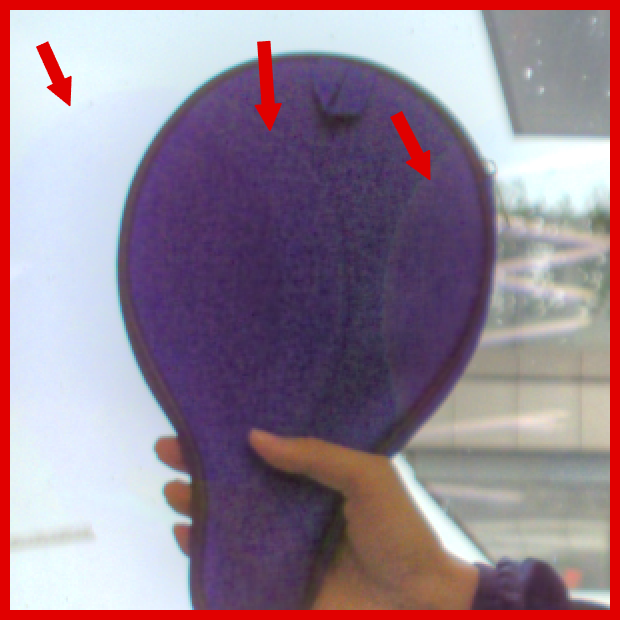}
		\end{subfigure}	
		\begin{subfigure}{\wseven\textwidth}
			\includegraphics[height=\hseven]{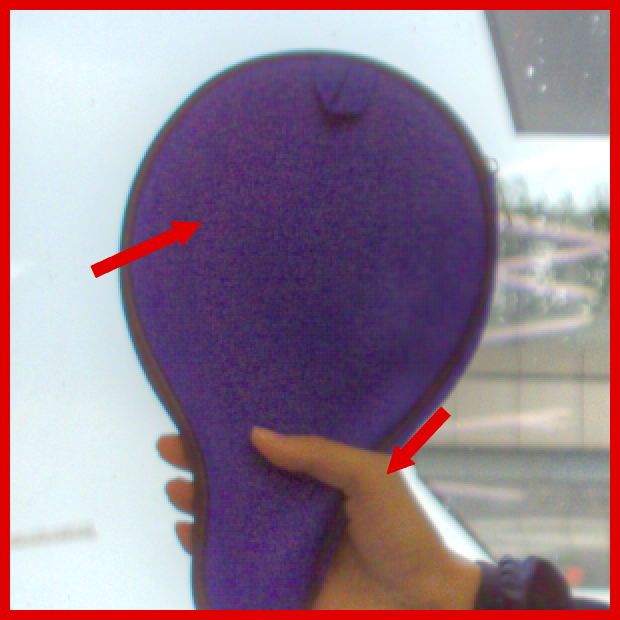}
		\end{subfigure}	
		\begin{subfigure}{\wseven\textwidth}
			\includegraphics[height=\hseven]{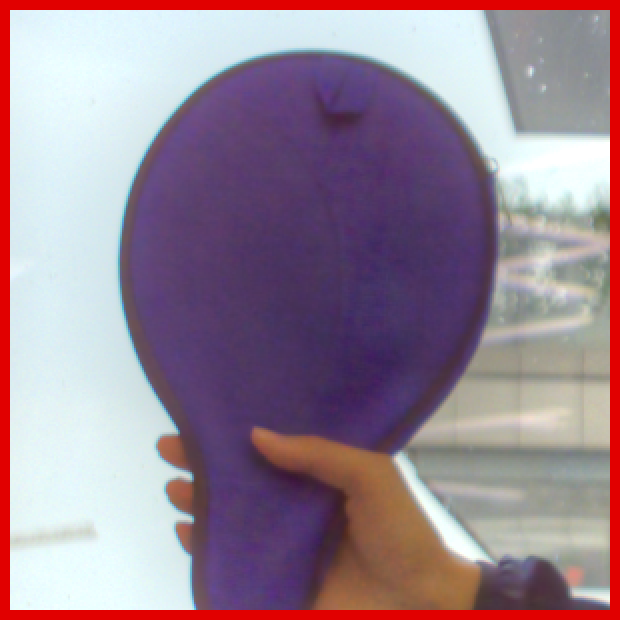}
		\end{subfigure}	
		\begin{subfigure}{0.0696\textwidth}
			\includegraphics[height=\hseven]{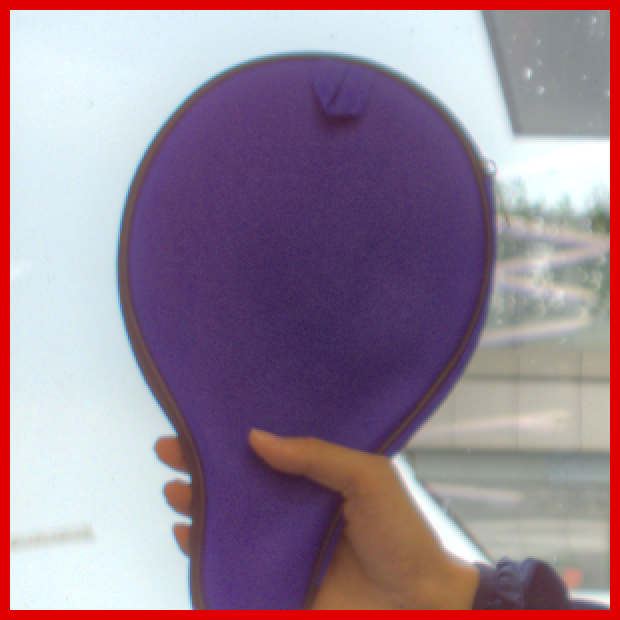}
		\end{subfigure}	
		\begin{subfigure}{\wseven\textwidth}
			\includegraphics[height=\hseven]{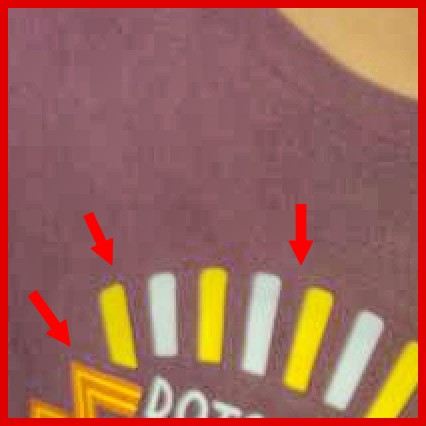}
		\end{subfigure}	
		\begin{subfigure}{\wseven\textwidth}
			\includegraphics[height=\hseven]{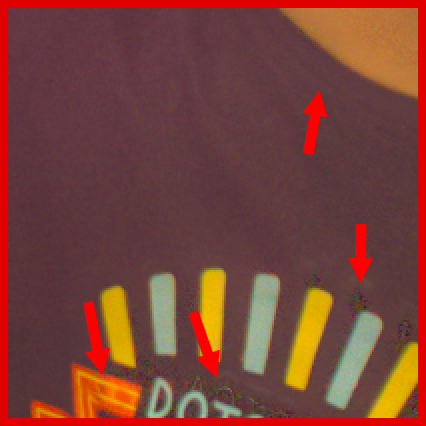}
		\end{subfigure}	
		\begin{subfigure}{\wseven\textwidth}
			\includegraphics[height=\hseven]{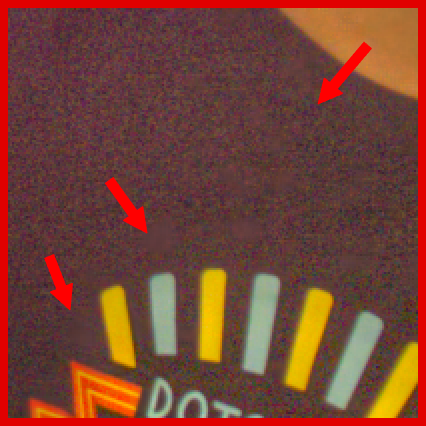}
		\end{subfigure}	
		\begin{subfigure}{\wseven\textwidth}
			\includegraphics[height=\hseven]{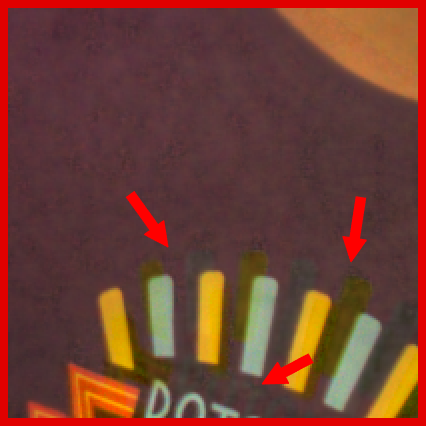}
		\end{subfigure}	
		\begin{subfigure}{\wseven\textwidth}
			\includegraphics[height=\hseven]{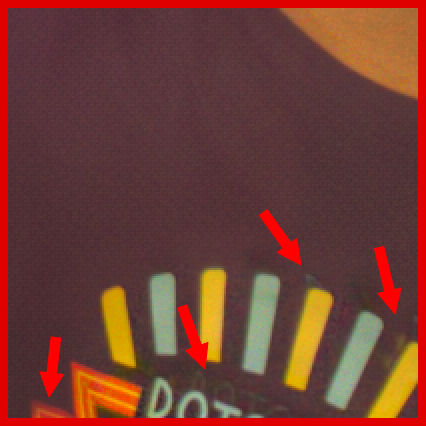}
		\end{subfigure}	
		\begin{subfigure}{\wseven\textwidth}
			\includegraphics[height=\hseven]{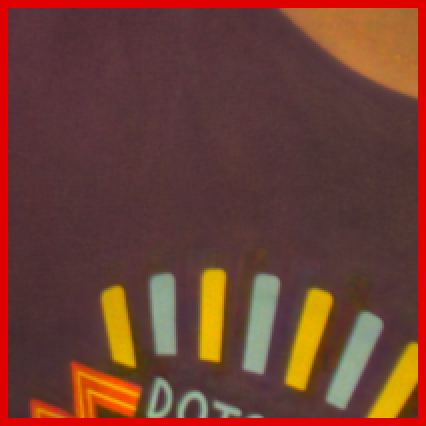}
		\end{subfigure}	
		\begin{subfigure}{\wsevenlast\textwidth}
			\includegraphics[height=\hseven]{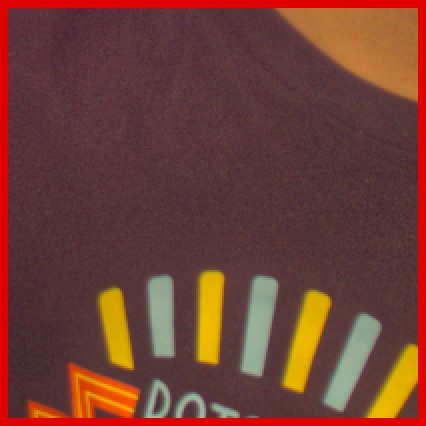}
		\end{subfigure}	
		
		\begin{subfigure}{\wseven\textwidth}
			\includegraphics[height=\hseven]{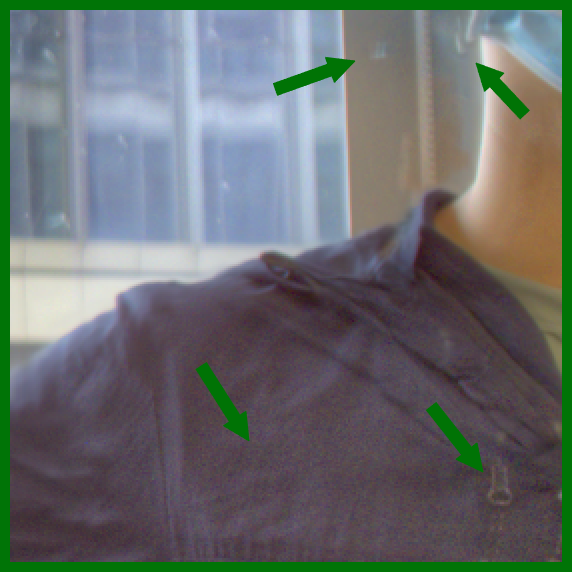}\vskip \mcaption
			\caption*{\footnotesize Kalantari} \vspace*{\rowmargin}
			\caption*{\footnotesize \cite{kalantari2013patch}}\vspace*{\bigrowmargin}
		\end{subfigure}	
		\begin{subfigure}{\wseven\textwidth}
			\includegraphics[height=\hseven]{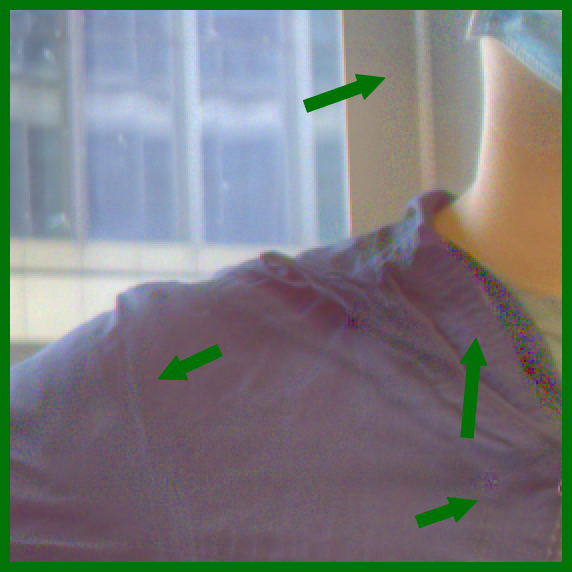} \vskip \mcaption
			\caption*{\footnotesize Kalantari} \vspace*{\rowmargin}
			\caption*{\footnotesize \cite{kalantari2019deep}}\vspace*{\bigrowmargin}
		\end{subfigure}	
		\begin{subfigure}{\wseven\textwidth}
			\includegraphics[height=\hseven]{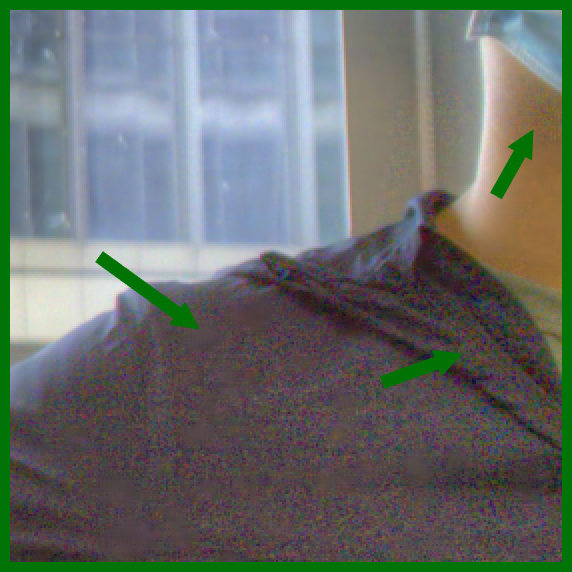}\vskip \mcaption
			\caption*{\footnotesize Yan} \vspace*{\rowmargin}
			\caption*{\footnotesize \cite{yan2019attention}}\vspace*{\bigrowmargin}
		\end{subfigure}	
		\begin{subfigure}{\wseven\textwidth}
			\includegraphics[height=\hseven]{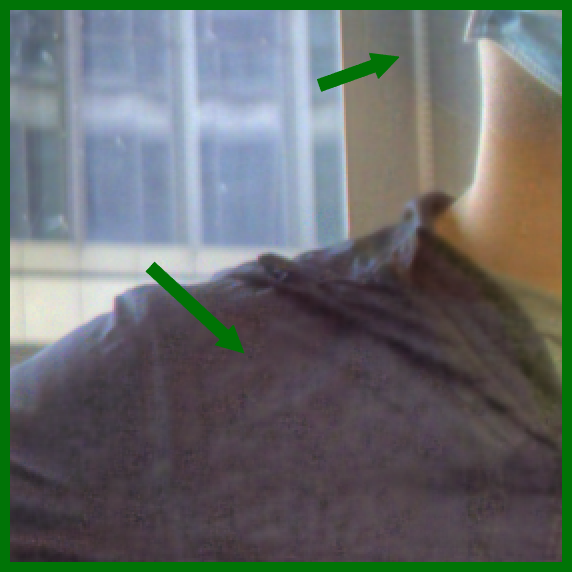} \vskip \mcaption
			\caption*{\footnotesize Prabhakar} \vspace*{\rowmargin}
			\caption*{\footnotesize \cite{prabhakar2021labeled}}\vspace*{\bigrowmargin}
		\end{subfigure}	
		\begin{subfigure}{\wseven\textwidth}
			\includegraphics[height=\hseven]{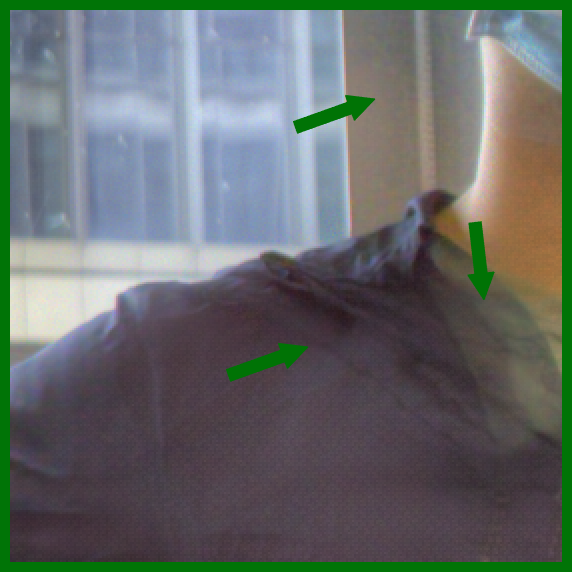}\vskip \mcaption
			\caption*{\footnotesize Chen} \vspace*{\rowmargin}
			\caption*{\footnotesize \cite{chen2021hdr}}\vspace*{\bigrowmargin}
		\end{subfigure}	
		\begin{subfigure}{\wseven\textwidth}
			\includegraphics[height=\hseven]{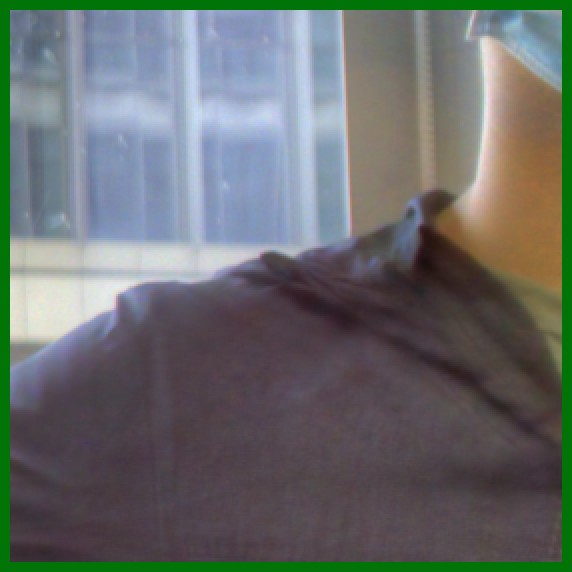}\vskip \mcaption
			\caption*{\footnotesize Ours} \vspace*{\rowmargin}
			\caption*{\footnotesize  }\vspace*{\bigrowmargin}
		\end{subfigure}	
		\begin{subfigure}{0.0696\textwidth}
			\includegraphics[height=\hseven]{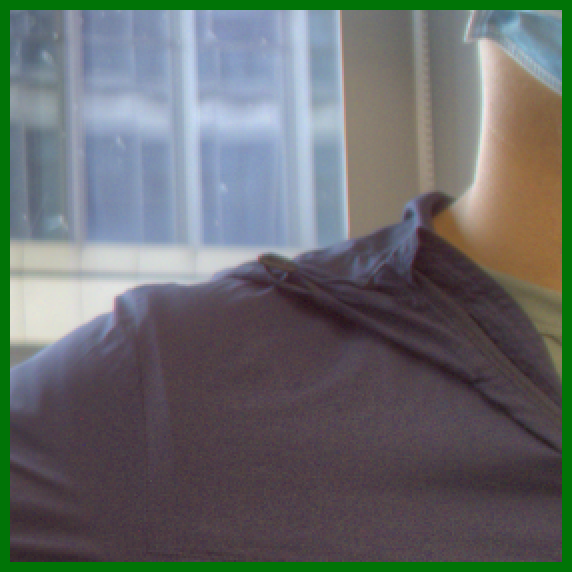} \vskip \mcaption
			\caption*{\footnotesize \! GT} \vspace*{\rowmargin}
			\caption*{\footnotesize }\vspace*{\bigrowmargin}
		\end{subfigure}	
		\begin{subfigure}{\wseven\textwidth}
			\includegraphics[height=\hseven]{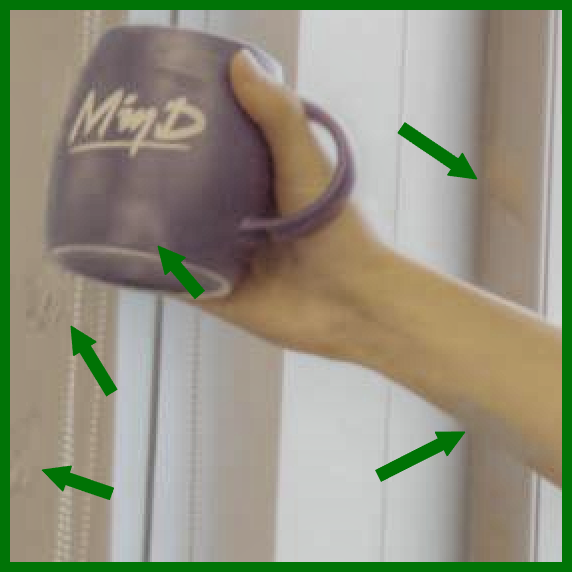}\vskip \mcaption
			\caption*{\footnotesize Kalantari} \vspace*{\rowmargin}
			\caption*{\footnotesize \cite{kalantari2013patch}}\vspace*{\bigrowmargin}
		\end{subfigure}	
		\begin{subfigure}{\wseven\textwidth}
			\includegraphics[height=\hseven]{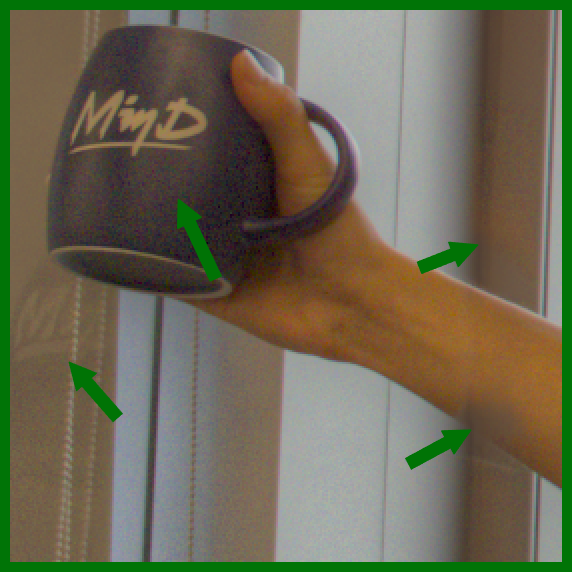}\vskip \mcaption
			\caption*{\footnotesize Kalantari} \vspace*{\rowmargin}
			\caption*{\footnotesize \cite{kalantari2019deep}}\vspace*{\bigrowmargin}
		\end{subfigure}	
		\begin{subfigure}{\wseven\textwidth}
			\includegraphics[height=\hseven]{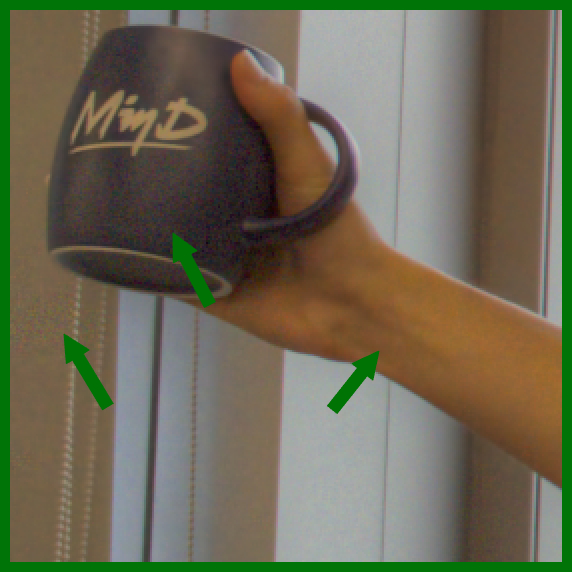}\vskip \mcaption
			\caption*{\footnotesize Yan} \vspace*{\rowmargin}
			\caption*{\footnotesize \cite{yan2019attention}}\vspace*{\bigrowmargin}
		\end{subfigure}	
		\begin{subfigure}{\wseven\textwidth}
			\includegraphics[height=\hseven]{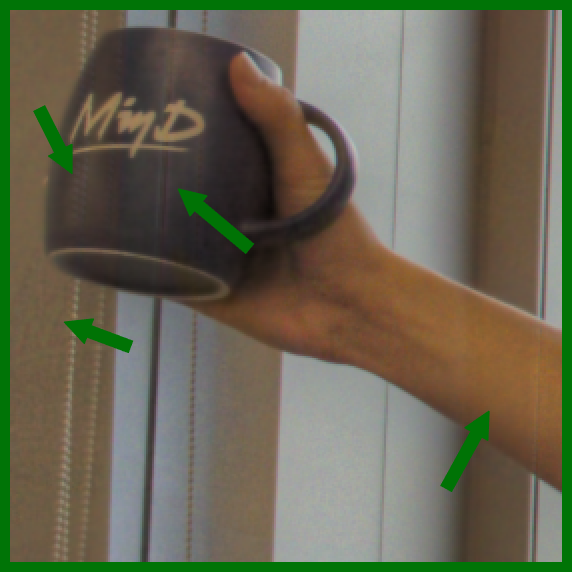} \vskip \mcaption
			\caption*{\footnotesize Prabhakar} \vspace*{\rowmargin}
			\caption*{\footnotesize \cite{prabhakar2021labeled}}\vspace*{\bigrowmargin}
		\end{subfigure}	
		\begin{subfigure}{\wseven\textwidth}
			\includegraphics[height=\hseven]{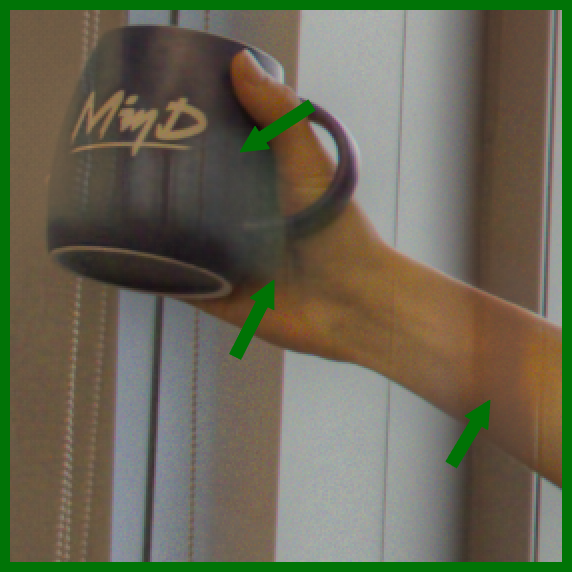} \vskip \mcaption
			\caption*{\footnotesize \! Chen} \vspace*{\rowmargin}
			\caption*{\footnotesize \cite{chen2021hdr}}\vspace*{\bigrowmargin}
		\end{subfigure}	
		\begin{subfigure}{\wseven\textwidth}
			\includegraphics[height=\hseven]{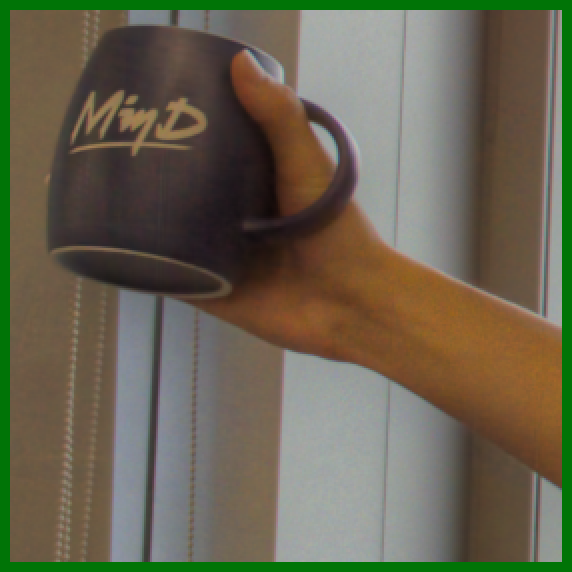} \vskip \mcaption
			\caption*{\footnotesize Ours} \vspace*{\rowmargin}
			\caption*{\footnotesize }\vspace*{\bigrowmargin}
		\end{subfigure}	
		\begin{subfigure}{\wsevenlast\textwidth}
			\includegraphics[height=\hseven]{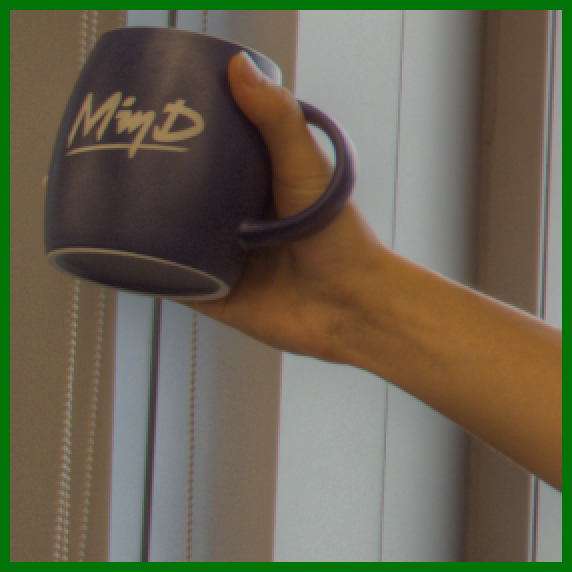}\vskip \mcaption
			\caption*{\footnotesize \quad \: GT} \vspace*{\rowmargin}
			\caption*{\footnotesize }   \vspace*{\bigrowmargin}
		\end{subfigure}	
		\caption{Qualitative results on $3$-exposure sequences in the DeepHDRVideo dataset \cite{chen2021hdr}. Only the middle three input frames are displayed.}
		\label{fig:3E} \vskip -1.05mm
	\end{figure*}

	\section{Experiments}
	\subsection{Experimental Setup}
	\noindent\textbf{Datasets} 
	We use synthetic training data constructed using the Vimeo-90K septuplet dataset \cite{xue2019video}. Since the Vimeo-90K dataset is not optimized for HDR video reconstruction, we convert the original data to LDR sequences with alternating exposures following the previous work~\cite{chen2021hdr}. We evaluate our framework on two synthetic videos (POKER FULLSHOT and CAROUSEL FIREWORKS) from the Cinematic Video dataset~\cite{froehlich2014creating} and DeepHDRVideo dataset \cite{chen2021hdr}. The DeepHDRVideo dataset~\cite{chen2021hdr} includes both real-world dynamic scenes and static scenes augmented with random global motion. HDRVideo dataset \cite{kalantari2013patch} is used only for qualitative evaluation due to the absence of ground truth.
	
	\noindent\textbf{Training details}
	We adopt AdamW optimizer~\cite{loshchilov2017decoupled} with $\beta_1 = 0.9$ and $\beta_2 = 0.999$, and the learning rate is set to $10^{-4}$. We augment training patches using random flipping, rotation by multiples of $90^{\circ}$, and color augmentation. More details are included in the supplementary material.
	
	\noindent\textbf{Evaluation metrics} 
	We compute PSNR\textsubscript{$T$}, SSIM\textsubscript{$T$}, PSNR\textsubscript{$PU$}, SSIM\textsubscript{$PU$}, and HDR-VDP-2 between the predicted results and ground truth frames. PSNR\textsubscript{$T$} and SSIM\textsubscript{$T$} are computed on the tonemapped images using the $\mu$-law. PSNR\textsubscript{$PU$} and SSIM\textsubscript{$PU$} are calculated after perceptually uniform encoding \cite{mantiuk2021pu21}, which maps the peak value in the HDR image to the peak HDR display luminance $4000 cd/m^2$. When computing the HDR-VDP-2, the angular resolution of the image in terms of the number of pixels per visual degree is set as $30$.
	
	\newcommand{\htog}{1.65cm}
	\newcommand{\wtog}{0.12}
	\newcommand{\togmargin}{0.05mm}
	
	\begin{figure*}[t]
		\begin{minipage}{.228\textwidth}
			\begin{subfigure}{.99\textwidth}
				\includegraphics[height=3.32cm]{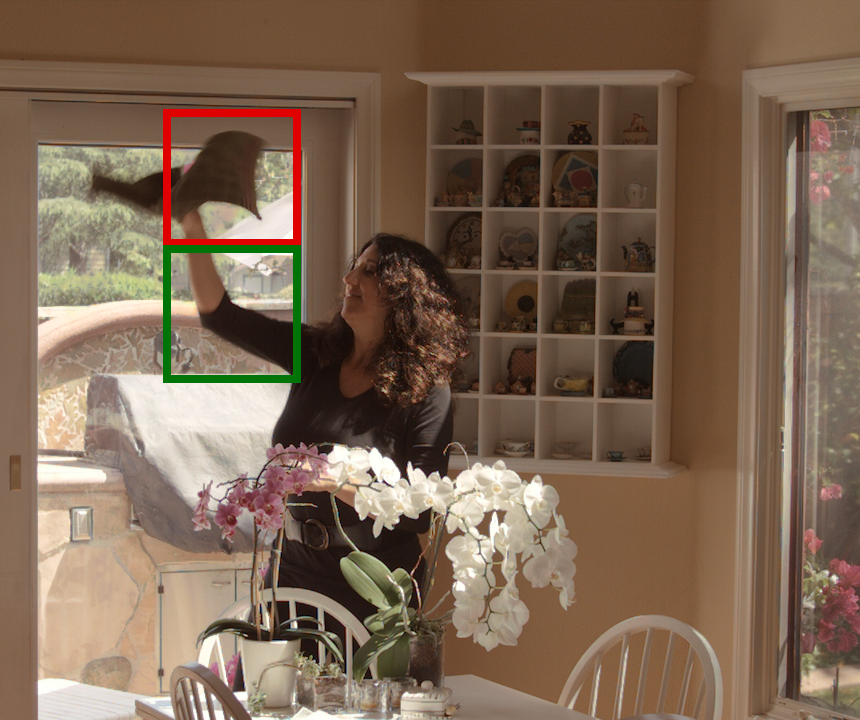}\vskip -0.6mm
				\caption*{\footnotesize Our Result} 
			\end{subfigure}	
		\end{minipage}
		\begin{minipage}{.78\textwidth}	
			\begin{subfigure}{\wtog\textwidth}
				\includegraphics[height=\htog]{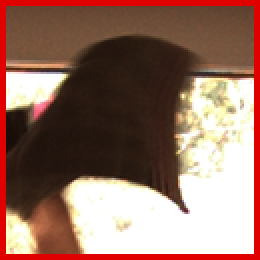} 
			\end{subfigure}			
			\begin{subfigure}{\wtog\textwidth}
				\includegraphics[height=\htog]{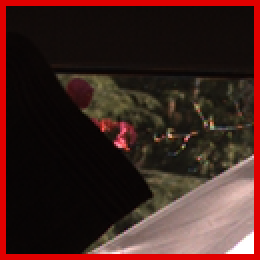} 
			\end{subfigure}			
			\begin{subfigure}{\wtog\textwidth}
				\includegraphics[height=\htog]{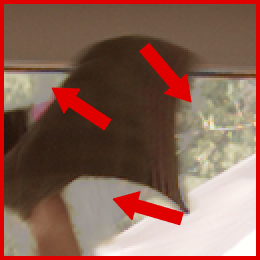} 
			\end{subfigure}	
			\begin{subfigure}{\wtog\textwidth}
				\includegraphics[height=\htog]{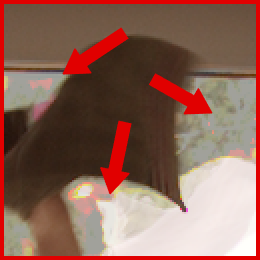} 
			\end{subfigure}	
			\begin{subfigure}{\wtog\textwidth}
				\includegraphics[height=\htog]{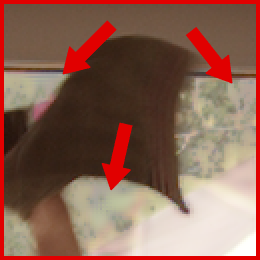} 
			\end{subfigure}	
			\begin{subfigure}{\wtog\textwidth}
				\includegraphics[height=\htog]{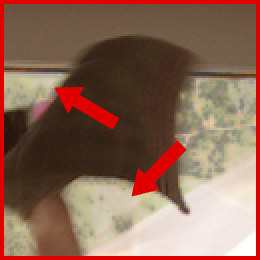} 
			\end{subfigure}
			\begin{subfigure}{\wtog\textwidth}
				\includegraphics[height=\htog]{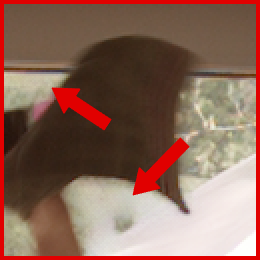} 
			\end{subfigure}	
			\begin{subfigure}{\wtog\textwidth}
				\includegraphics[height=\htog]{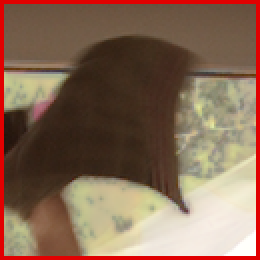} 
			\end{subfigure}
			
			\begin{subfigure}{\wtog\textwidth}
				\includegraphics[height=\htog]{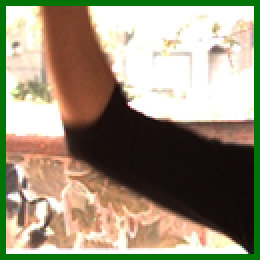} \vskip \mcaption
				\caption*{\footnotesize \qquad \quad \:\: Input} 
			\end{subfigure}	\vspace{\togmargin}	
			\begin{subfigure}{\wtog\textwidth}
				\includegraphics[height=\htog]{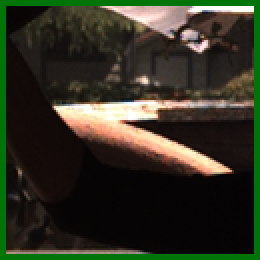} \vskip \mcaption
				\caption*{\footnotesize Frames \:\:\qquad\qquad} 
			\end{subfigure}	\vspace{\togmargin}		
			\begin{subfigure}{\wtog\textwidth}
				\includegraphics[height=\htog]{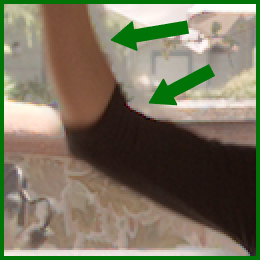} \vskip \mcaption
				\caption*{\footnotesize Kalantari \cite{kalantari2013patch}} 
			\end{subfigure}	\vspace{\togmargin}
			\begin{subfigure}{\wtog\textwidth}
				\includegraphics[height=\htog]{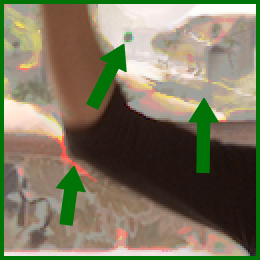} \vskip \mcaption
				\caption*{\footnotesize \! Kalantari \cite{kalantari2019deep}} 
			\end{subfigure}\vspace{\togmargin}	
			\begin{subfigure}{\wtog\textwidth}
				\includegraphics[height=\htog]{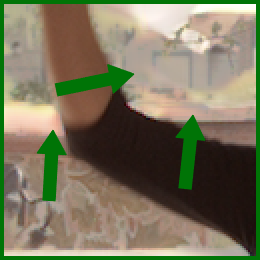} \vskip \mcaption
				\caption*{\footnotesize Yan \cite{yan2019attention}} 
			\end{subfigure}	\vspace{\togmargin}
			\begin{subfigure}{\wtog\textwidth}
				\includegraphics[height=\htog]{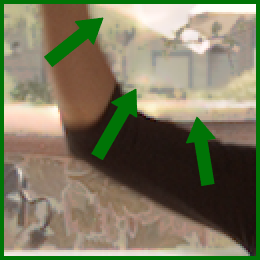}\vskip \mcaption
				\caption*{\footnotesize \! Prabhakar \scriptsize\cite{prabhakar2021labeled}} 	\vspace{0.16mm}	
			\end{subfigure}	\vspace{\togmargin}		
			\begin{subfigure}{\wtog\textwidth}
				\includegraphics[height=\htog]{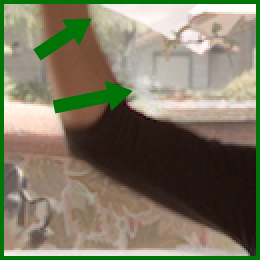} \vskip \mcaption
				\caption*{\footnotesize \: Chen \cite{chen2021hdr}} 
			\end{subfigure}	\vspace{\togmargin}
			\begin{subfigure}{\wtog\textwidth}
				\includegraphics[height=\htog]{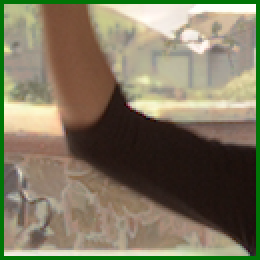} \vskip \mcaption
				\caption*{\footnotesize Ours} 
			\end{subfigure}\vspace{\togmargin}
		\end{minipage}
		\caption{Qualitative comparisons on the sequence with $2$ exposures from the HDRVideo dataset \cite{kalantari2013patch}. Only the middle two input frames are displayed.}
		\label{fig:2E}
	\end{figure*}

	\subsection{Comparisons}
	We compare our results with previous HDR video reconstruction approaches \cite{kalantari2013patch, kalantari2019deep, chen2021hdr} and state-of-the-art HDR image composition methods \cite{yan2019attention, prabhakar2021labeled} after adapting them to the video reconstruction task. We use the official codes if available, otherwise, we re-implement them based on the papers. For a fair comparison, we have trained deep learning-based methods with the dataset we use. All the visual results are tonemapped using Photomatix~\cite{Photomatix} and best viewed by zooming into the electronic version.
	
	\noindent\textbf{Quantitative evaluations} 
	Our results on Cinematic Video dataset~\cite{froehlich2014creating} and DeepHDRVideo dataset~\cite{chen2021hdr} are listed in Table~\ref{tab:synthetic} and Table~\ref{tab:real}, respectively, where our approach achieves superior or comparable performances to state-of-the-art methods. Especially, in the case of data with $3$ exposures, our method consistently demonstrates precise reconstruction ability by handling large variations of motions and brightness. 
	
	\noindent\textbf{Qualitative evaluations} 
	We visualize our results on sequences with saturation and large motions. Fig.~\ref{fig:3E} shows the results on two $3$-exposure sequences from the DeepHDRVideo dataset \cite{chen2021hdr}. The other methods except ours fail to avoid ghosting artifacts and color distortions when fast motion occurs with exposure changes. Especially, the approaches using flow-based estimation \cite{kalantari2019deep, chen2021hdr} suffer from artifacts caused by flow estimation error. We also evaluate on a sequence with two alternating exposures from the HDRVideo dataset~\cite{kalantari2013patch}. Fig.~\ref{fig:2E} shows that our method generates fine details even when large occlusion and saturation exist. More visual results are presented in Fig.~\ref{fig:teaser} and the supplementary material.
	
	\newcommand{\htemporals}{0.88cm}
	\newcommand{\htemporal}{0.88}
	\newcommand{\mtemporal}{-1.0mm}
	\newcommand{\captionspace}{0.2mm}
	\begin{figure}[t]
		\begin{minipage}{0.105\textwidth}
			\begin{subfigure}{0.98\textwidth}
				\includegraphics[height=4.62cm]{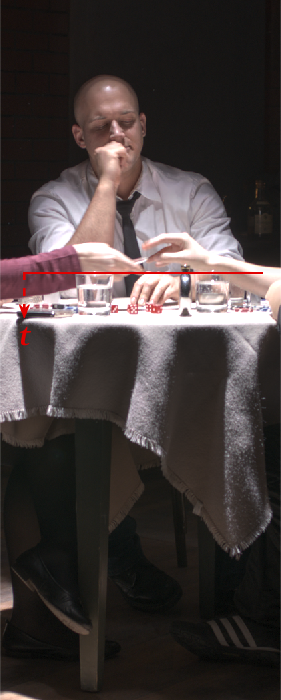} 
			\end{subfigure}
			\vskip \rowmargin	 \caption*{\small GT} 
		\end{minipage}
		\begin{minipage}{0.39\textwidth}
			\begin{subfigure}{\htemporal\textwidth}
				\includegraphics[height=\htemporals]{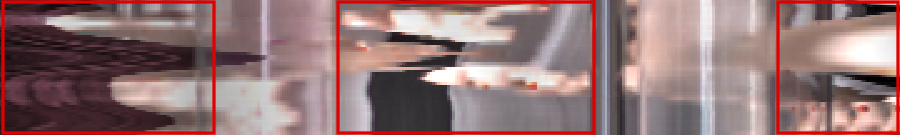} 
			\end{subfigure}
			\hspace*{\mtemporal}\rotatebox{90}{ \small \! \! \! \cite{kalantari2013patch} } 	
			\vspace*{\captionspace}\newline 
			\begin{subfigure}{\htemporal\textwidth}
				\includegraphics[height=\htemporals]{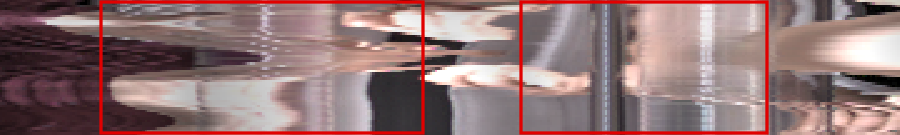}
			\end{subfigure}	
			\hspace*{\mtemporal}\rotatebox{90}{ \small \! \! \cite{kalantari2019deep} } 	
			\vspace*{\captionspace}\newline 
			\begin{subfigure}{\htemporal\textwidth}
				\includegraphics[height=\htemporals]{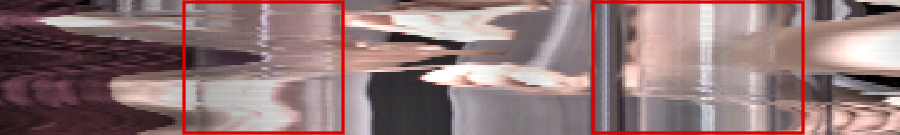}
			\end{subfigure}
			\hspace*{\mtemporal}\rotatebox{90}{ \small \: \cite{chen2021hdr} } 		
			\vspace*{\captionspace}\newline 
			\begin{subfigure}{\htemporal\textwidth}
				\includegraphics[height=\htemporals]{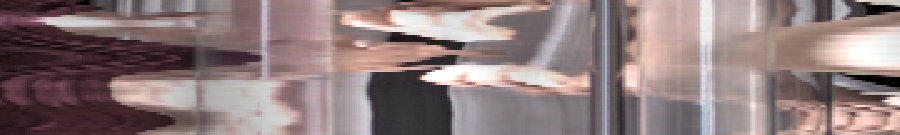}
			\end{subfigure}
			\hspace*{\mtemporal}\rotatebox{90}{ \small Ours } 	
			\vspace*{\captionspace}\newline 
			\begin{subfigure}{\htemporal\textwidth}
				\includegraphics[height=\htemporals]{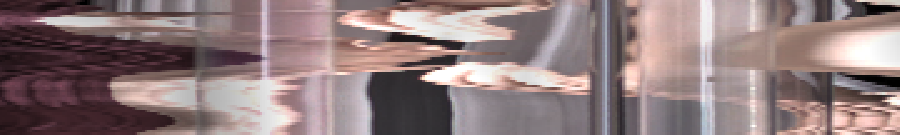} 	
			\end{subfigure}
			\hspace*{\mtemporal}\rotatebox{90}{ \small \:  GT } 
			\vskip \rowmargin	
			\caption*{\small Temporal Profiles \qquad\qquad}
		\end{minipage}
		
		\caption{Visualization of temporal consistency.} 
		\label{fig:3e_temporal}
	\end{figure}

	\begin{table}[t]
		\vskip -0.3mm
		\caption{Comparisons of inference time.}
		\centering
		\label{tab:time}
		\resizebox{\columnwidth}{!}{
			\begin{tabular}{l|cc|cc}
				\hline
				& \multicolumn{2}{c|}{2 Exposures}      & \multicolumn{2}{c}{3 Exposures}       \\ \cline{2-5} 
				& $1280 \times 720$ & $1536 \times 813$ & $1280 \times 720$ & $1536 \times 813$ \\ \hline
				Kalantari \cite{kalantari2013patch} & 140.02s            & 166.10s            & 321.16s            & 414.44s            \\
				Kalantari \cite{kalantari2019deep}  & 0.37s              & 0.69s              & 0.49s              & 0.72s              \\
				Chen \cite{chen2021hdr}      		& 0.54s              & 1.25s              & 0.54s              & 1.30s              \\
				Ours      							& 0.37s              & 0.61s              & 0.57s              & 0.82s              \\ \hline
			\end{tabular}
		}
	\end{table}
	
	\noindent\textbf{Temporal profiles} 
	This section provides temporal profiles to demonstrate the temporal consistency of our results. We produce temporal profiles by recording a single line across all the frames and stitching them in order. Fig.~\ref{fig:3e_temporal} shows the results on a $3$-exposure sequence in the Cinematic Video dataset \cite{froehlich2014creating}. Our method generates temporally consistent videos with minimal flickering artifacts and preserves detailed texture successfully.
	
	
	\noindent\textbf{Computation time} 
	We compare the inference time of our method with other HDR video construction methods in Table~\ref{tab:time}. We use a single NVIDIA TITAN Xp GPU, while the optimization-based method of Kalantari \etal~\cite{kalantari2013patch} is run on CPUs. Our approach is faster than the method of Chen \etal~\cite{chen2021hdr}, and the gap increases as the image resolution becomes larger.
	
	
	\begin{table}[t]
		\caption{Effectiveness of each component in the LAN. AM, HM, and ABL denote the alignment module, the hallucination module, and the adaptive blending layer, respectively.}
		\centering
		\label{tab:lan}
		\resizebox{0.95\columnwidth}{!}{
			\begin{tabular}{l|ccc|cc}
				\hline
				Model & AM  & HM & ABL& PSNR\textsubscript{$T$}  & PSNR\textsubscript{$PU$} \\ \hline 
				(a) & \checkmark &    &                		 & 36.37  & 37.99  \\
				(b) & & \checkmark  &        		 &  36.32 & 36.68  \\
				(c) & \checkmark    & \checkmark  &     		& 37.57  & 38.96 \\
				(d) & \checkmark   & \checkmark  & \checkmark  & \textbf{38.22} & \textbf{40.04} \\ \hline
			\end{tabular}
		}
	\end{table}

	\newcommand{\hfour}{2.05cm}
	\newcommand{\wfour}{.116}
	\newcommand{\m}{-0.5mm}

	\begin{figure}[t]
		\begin{subfigure}{\wfour\textwidth}
			\includegraphics[height=\hfour]{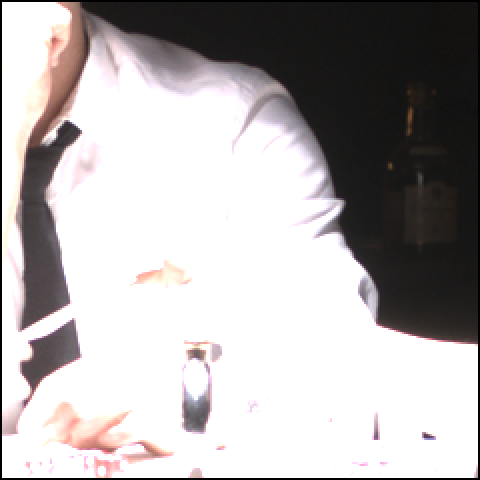}\vskip \mcaption
			\caption*{\footnotesize $L_{t}$} \vspace*{0.5mm}
		\end{subfigure}	
		\begin{subfigure}{\wfour\textwidth}
			\includegraphics[height=\hfour]{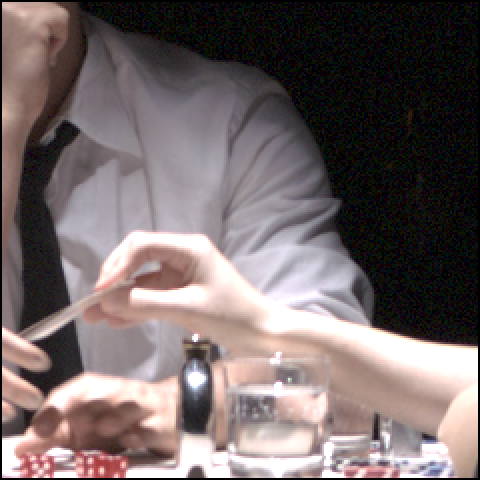}\vskip \mcaption
			\caption*{\footnotesize $L_{t+1}$} \vspace*{0.5mm}
		\end{subfigure}	
		\begin{subfigure}{\wfour\textwidth}
			\includegraphics[height=\hfour]{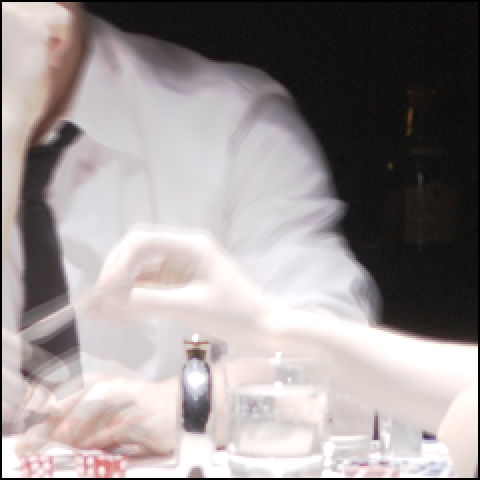}\vskip \mcaption
			\caption*{\footnotesize  Overlapped Input}\vspace*{0.5mm}
		\end{subfigure}	
		\begin{subfigure}{\wfour\textwidth}
			\includegraphics[height=\hfour]{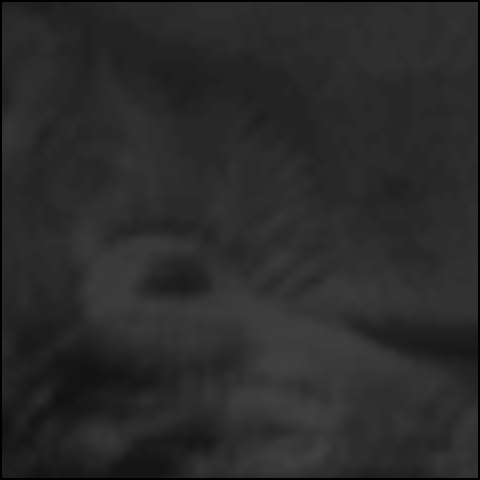} \vskip \mcaption
			\caption*{\footnotesize \! \! \! Blending Map $M$}\vspace*{0.5mm}
		\end{subfigure}
		\begin{subfigure}{\wfour\textwidth}
			\includegraphics[height=\hfour]{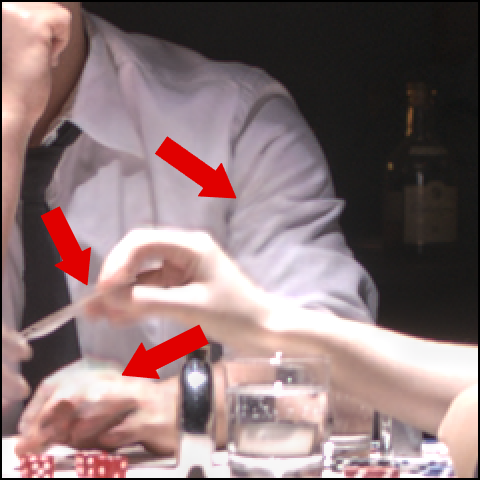}\vskip \mcaption
			\caption{\footnotesize AM Only} \vspace*{\bigrowmargin}\label{fig:lan_amonly}
		\end{subfigure}	
		\begin{subfigure}{\wfour\textwidth}
			\includegraphics[height=\hfour]{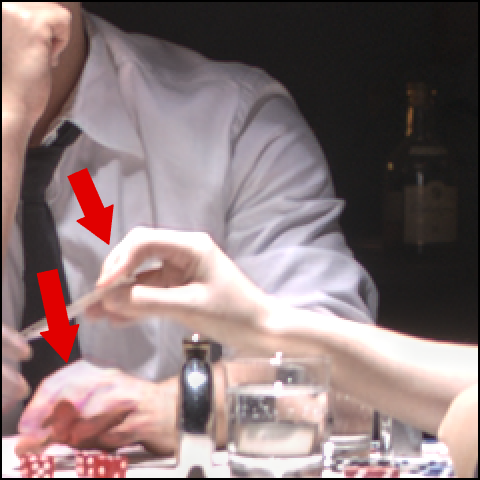}\vskip \mcaption
			\caption{\footnotesize HM Only} \vspace*{\bigrowmargin}\label{fig:lan_hmonly}
		\end{subfigure}	
		\begin{subfigure}{\wfour\textwidth}
			\includegraphics[height=\hfour]{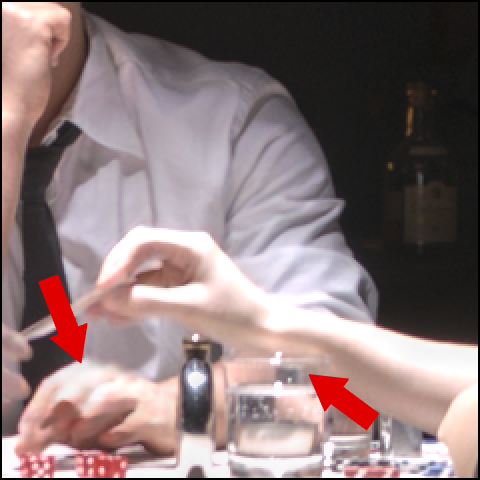}\vskip \mcaption
			\caption{\footnotesize w/o ABL} \vspace*{\bigrowmargin}\label{fig:lan_woabl}
		\end{subfigure}	
		\begin{subfigure}{\wfour\textwidth}
			\includegraphics[height=\hfour]{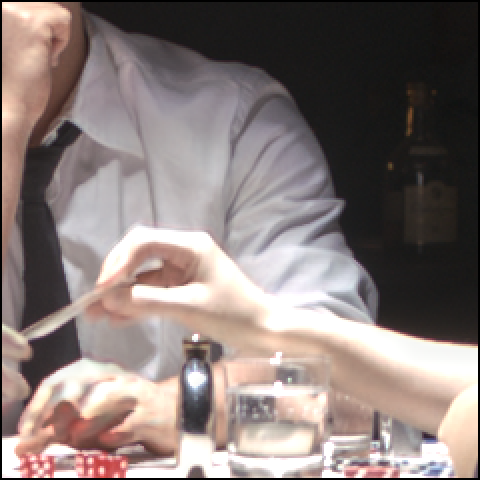}\vskip \mcaption
			\caption{\footnotesize Ours} \vspace*{\bigrowmargin}\label{fig:lan_ours}
		\end{subfigure}	
		\caption{Analysis of the components in the LAN.}
		\label{fig:lan}
	\end{figure}
	
	\subsection{Analysis} \label{ablations}
	In this section, we analyze the effects of the proposed components with the example of sequences having two alternating exposures. All the quantitative evaluations are conducted on the Cinematic Video dataset \cite{froehlich2014creating}.
	
	\noindent\textbf{LAN} 
	Our LAN is composed of the alignment module (AM) and the hallucination module (HM), and features from the two modules are fused in the adaptive blending layer (ABL). As shown in Table~\ref{tab:lan}, the model with only AM or HM achieves far lower PSNRs than the full proposed model. The model which has both AM and HM, but blends two features using a simple concatenation, improves performance. But it suffers from undesirable artifacts in a challenging case such as a sequence with alternating exposures whose middle frame has large saturations as in Fig.~\ref{fig:lan}. 
	In this case, the AM cannot perform a perfect alignment because a large amount of content in the reference frame is totally washed out and thus produces blurry texture, as shown in Fig.~\ref{fig:lan_amonly}. On the other hand, Fig.~\ref{fig:lan_hmonly} shows that the HM generates sharp details but fails to align the motions according to the reference. (See the hand holding a card.) Our full model, including the AM, HM, and ABL (Fig.~\ref{fig:lan_ours}), outputs better results than the individual modules by properly combining two kinds of features. For example, our blending map has low values around the moving object in saturated parts, demonstrating that inaccurate AM features are suppressed in those areas. 

	\begin{figure}[t]
		\begin{subfigure}{\wfour\textwidth}
			\includegraphics[height=\hfour]{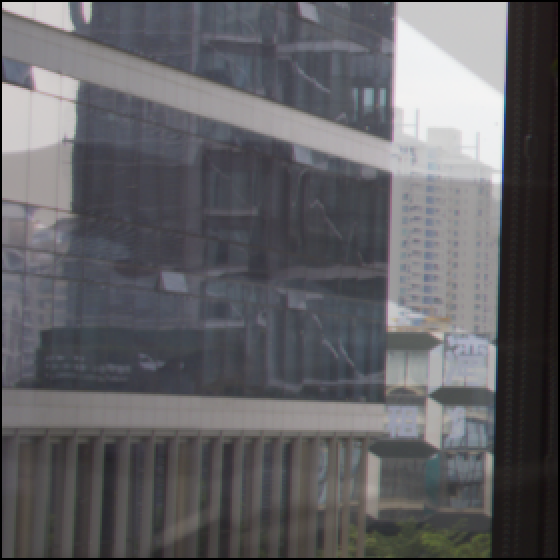}\vskip \mcaption
			\caption*{\scriptsize Overlapped Input} \vspace*{\bigrowmargin}
		\end{subfigure}	
		\begin{subfigure}{\wfour\textwidth}
			\includegraphics[height=\hfour]{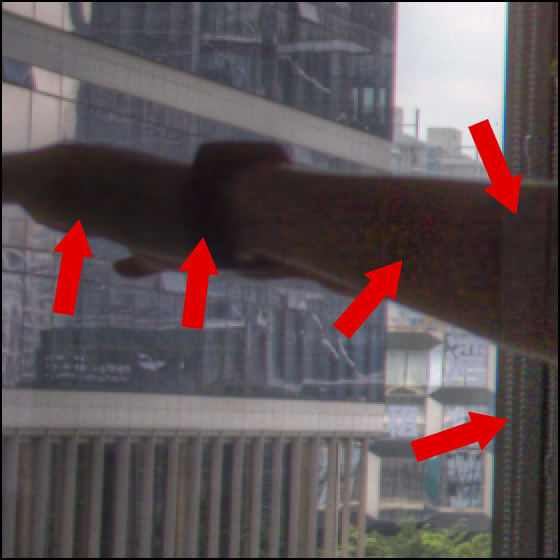}\vskip \mcaption
			\caption*{\scriptsize \! RGB Image} \vspace*{\bigrowmargin}
		\end{subfigure}	
		\begin{subfigure}{\wfour\textwidth}
			\includegraphics[height=\hfour]{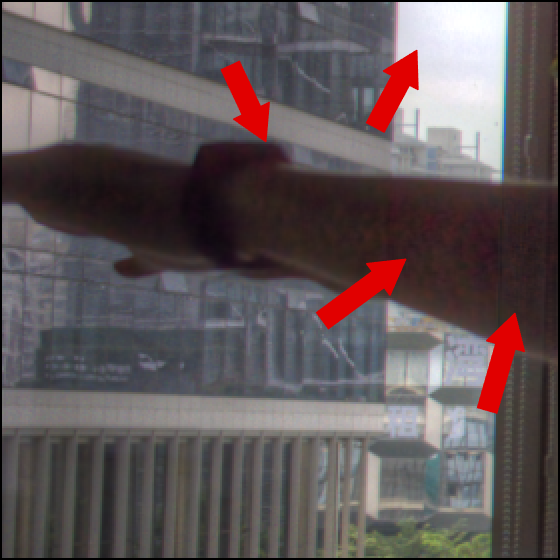}\vskip \mcaption
			\caption*{ \scriptsize Color Jittered Image} \vspace*{\bigrowmargin}
		\end{subfigure}	
		\begin{subfigure}{\wfour\textwidth}
			\includegraphics[height=\hfour]{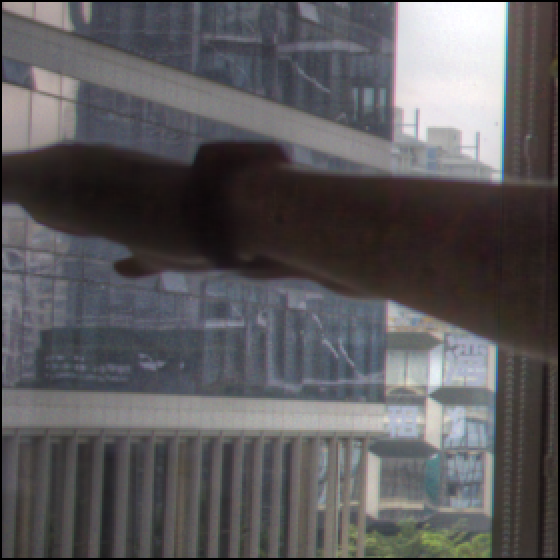}\vskip \mcaption
			\caption*{\scriptsize \! Y Channel (Ours)} \vspace*{\bigrowmargin}
		\end{subfigure}	
		\caption{Results with different inputs to the key/query extractor of the alignment module.}
		\label{fig:align}
	\end{figure}

	\begin{table}[t]
		\caption{Effectiveness of gated convolution and input masks in the hallucination module.}
		\centering
		\label{tab:hall}
		\resizebox{0.95\columnwidth}{!}{
			\begin{tabular}{l|cc|cc}
				\hline
				Model & \multicolumn{1}{c}{Convolution} & Input Mask & PSNR\textsubscript{$T$} & PSNR\textsubscript{$PU$}  \\ \hline
				(A) & Regular	& Y channel    & 37.02      & 38.55                     \\
				(B) & Gated  & -             & 36.49      & 38.10                    \\
				(C) & Gated  & Binary & 36.88     & 38.32                    \\
				(D) & Gated  & Y channel      & \textbf{38.22}  & \textbf{40.04}             \\ \hline
			\end{tabular}
		}\vskip -0.2mm
	\end{table}
	
	\begin{figure}[t!]
		\begin{subfigure}{\wfour\textwidth}
			\includegraphics[height=\hfour]{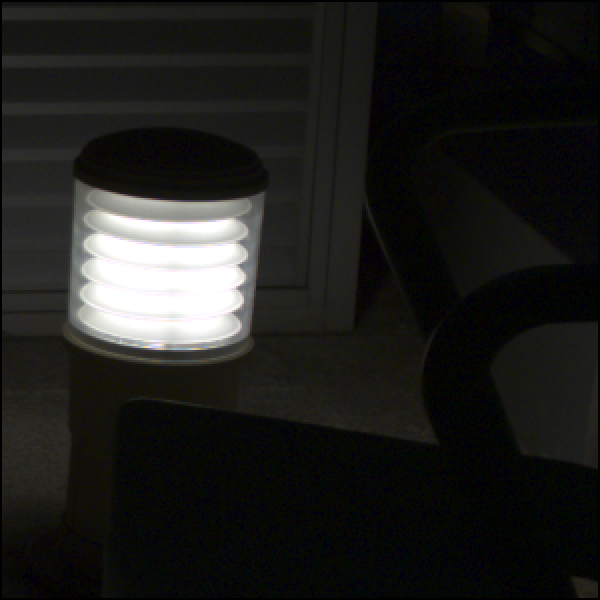} \vskip \mcaption
			\caption*{\footnotesize $L_{t}$}  \vspace*{0.5mm}
		\end{subfigure}	
		\begin{subfigure}{\wfour\textwidth}
			\includegraphics[height=\hfour]{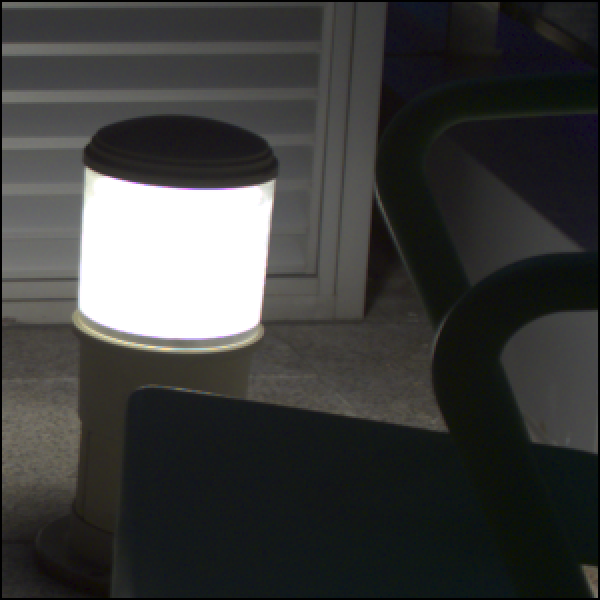}\vskip \mcaption
			\caption*{\footnotesize $L_{t+1}$}  \vspace*{0.5mm}
		\end{subfigure}	
		\begin{subfigure}{\wfour\textwidth}
			\includegraphics[height=\hfour]{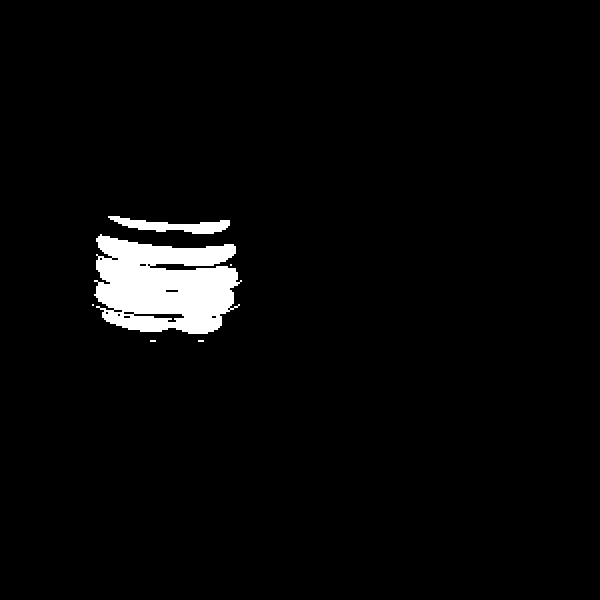}\vskip \mcaption
			\caption*{\footnotesize Mask of (C)}  \vspace*{0.5mm}
		\end{subfigure}	
		\begin{subfigure}{\wfour\textwidth}
			\includegraphics[height=\hfour]{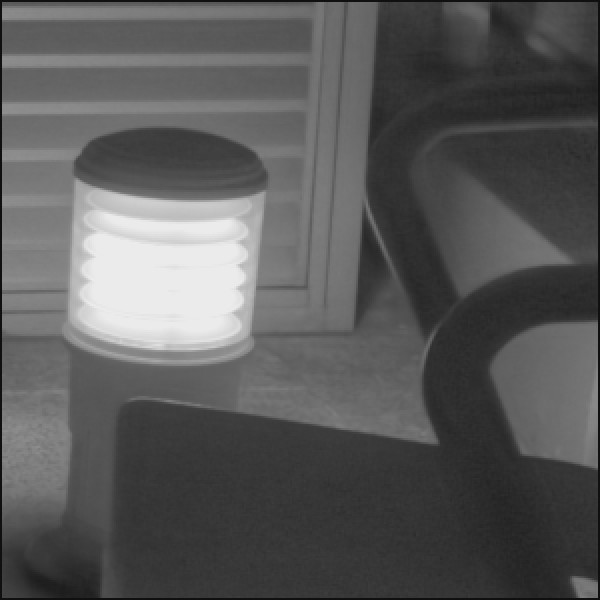} \vskip \mcaption
			\caption*{\footnotesize Mask of (D)} \vspace*{0.5mm}
		\end{subfigure}
		\begin{subfigure}{\wfour\textwidth}
			\includegraphics[height=\hfour]{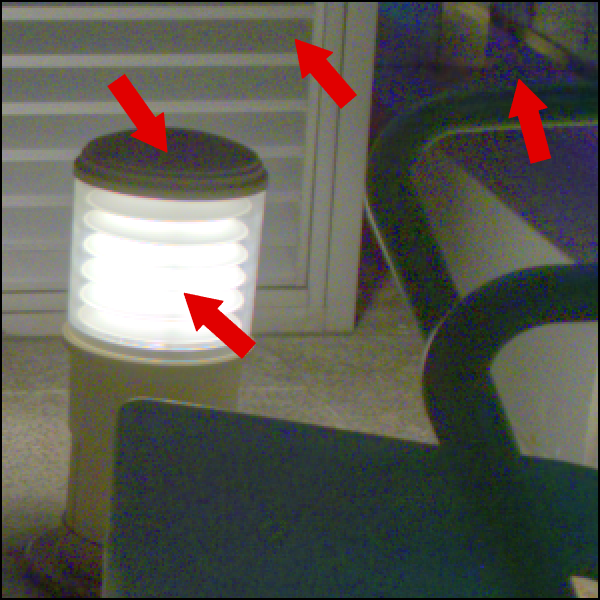}\vskip \mcaption
			\caption*{\footnotesize Model (A)} \vspace*{\bigrowmargin}
		\end{subfigure}	
		\begin{subfigure}{\wfour\textwidth}
			\includegraphics[height=\hfour]{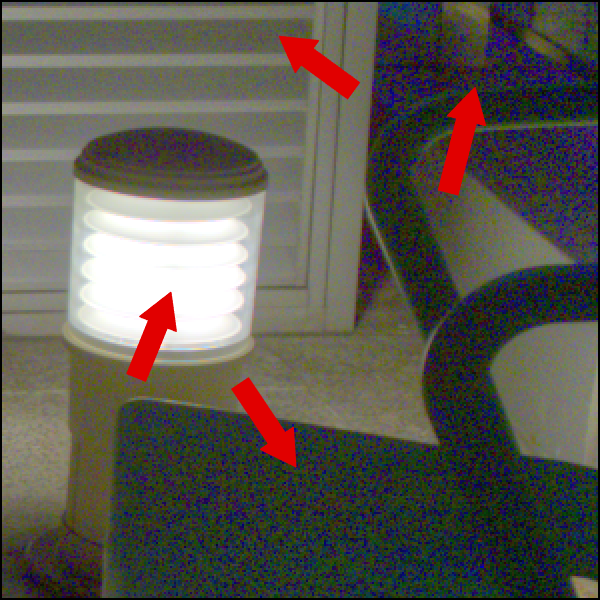}\vskip \mcaption
			\caption*{\footnotesize Model (B)} \vspace*{\bigrowmargin}
		\end{subfigure}	
		\begin{subfigure}{\wfour\textwidth}
			\includegraphics[height=\hfour]{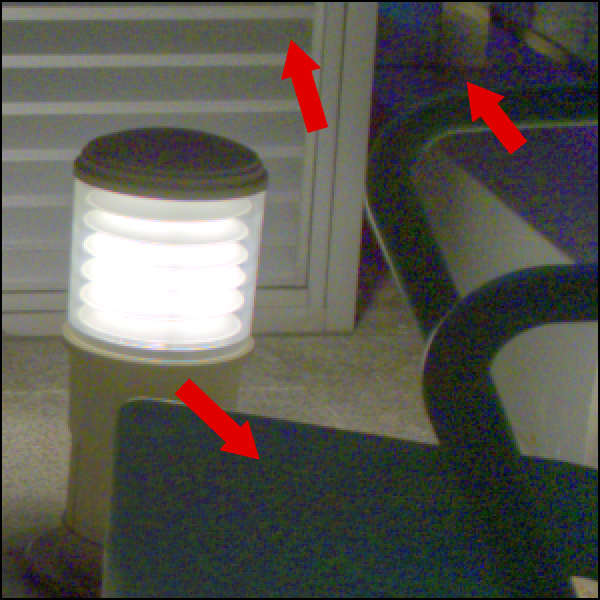}\vskip \mcaption
			\caption*{\footnotesize Model (C)} \vspace*{\bigrowmargin}
		\end{subfigure}	
		\begin{subfigure}{\wfour\textwidth}
			\includegraphics[height=\hfour]{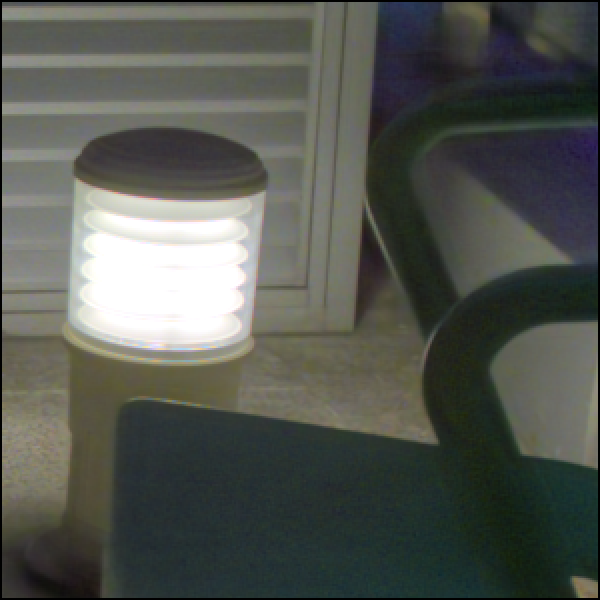}\vskip \mcaption
			\caption*{\footnotesize Model (D) (Ours)} \vspace*{\bigrowmargin}
		\end{subfigure}	
		\caption{Results with different convolutions and input masks in the hallucination module.}
		\label{fig:hall}\vskip -1.6mm
	\end{figure}

	\noindent\textbf{Inputs for alignment module}
	In our AM, we perform content-based alignment using Y channels of two input frames to prevent a simplistic matching based on color information. We compare our model with the one that uses original RGB images for alignment in Fig.~\ref{fig:align}. When using RGB images for attention operation, registration is unsuccessful when objects with similar colors overlap. We also conduct experiments with color-jittered RGB images as another means of relying less on color and more on content. The model using the color-jittered images improves alignment, but our model shows the best alignment results.

	\noindent\textbf{Hallucination module}
	Our HM in the LAN utilizes gated convolution and input masks to operate adaptively to luminance. The effects of these elements are shown in Table~\ref{tab:hall} and Fig.~\ref{fig:hall}. Model (A) with regular convolution takes the Y channel as input, but the dynamic operation is not possible. Model (B) includes gated convolution, but there is no explicit instruction for adaptive convolution, resulting in unsatisfying results. For model (C), a binary mask obtained from thresholding at a predefined value of $0.95$ is given. The binary mask indicates saturated parts but provides limited information regarding brightness. Fig.~\ref{fig:hall} shows a scene with both highlighted and very dark regions. It is noteworthy that other models not only lack details in bright parts but also generate significant noise in dark areas. In contrast, our framework suppresses noise in dark regions and produces detailed content in saturated regions adaptively.

	\begin{table}[t]
		\caption{Effectiveness of the proposed temporal loss.}
		\centering
		\label{tab:loss}
		\resizebox{0.95\columnwidth}{!}{
			\begin{tabular}{l|cccc}
				\hline
				Methods& PSNR\textsubscript{$T$} &SSIM\textsubscript{$T$}   & PSNR\textsubscript{$PU$} & SSIM\textsubscript{$PU$}  \\ \hline
				w/o $\mathcal{L}_{temp}$ & 36.33 & \textbf{0.9200}    & 37.99 & 0.8972               \\
				w/ $\mathcal{L}_{temp}$ & \textbf{38.22} & 0.9100 & \textbf{40.04} & \textbf{0.9039}           \\ \hline
			\end{tabular}
		}
	\end{table}
	
	\noindent\textbf{Temporal loss}
	We exploit the temporal loss $\mathcal{L}_{temp}$ to generate a temporally coherent HDR video. The temporal loss forces difference of consecutive frames to be similar to one of the ground truth. Table~\ref{tab:loss} shows that the overall quantitative performance is improved by using the temporal loss.

	\section{Conclusion}
	We have proposed an end-to-end HDR video composition framework using LDR videos with alternating exposures. To  enhance alignment accuracy and restore washed-out regions, our alignment network performs content-based alignment with attention and recovers details in saturated areas using brightness-adaptive convolution. The aligned features and the hallucinated features are fused adaptively using a learned blending map. The outputs of the alignment module are reconstructed into an HDR frame through the merging network. During training, the temporal loss is adopted for temporally consistent results. The overall results demonstrate that the proposed method generates high-quality HDR videos.
	
	\section*{Acknowledgments}
	
	This research was supported in part by Samsung Electronics Co., Ltd., in part by the National Research Foundation of Korea(NRF) grant funded by the Korea government(MSIT) (2021R1A2C2007220), and partially by Institute of Information \& communications Technology Planning \& Evaluation (IITP) grant funded by the Korea government(MSIT) (No. 2021-0-01062).
	
	%
	
	{\small
		\bibliographystyle{ieee_fullname}
		\bibliography{egbib}
	}
	
\end{document}